\documentclass[sigconf]{acmart}

\usepackage{url}
\usepackage{graphicx}
\usepackage{algorithmic}
\usepackage{subfigure}
\usepackage{amsmath}
\usepackage{xcolor}
\usepackage{makecell}
\usepackage{booktabs}
\usepackage{multirow}
\usepackage{balance}
\usepackage{listings}
\usepackage{booktabs}
\usepackage{pifont}
\usepackage{enumitem}
\usepackage[ruled,vlined,linesnumbered]{algorithm2e}
\usepackage[inkscapelatex=false]{svg}
\usepackage{bm}
\usepackage{array}
\usepackage{ragged2e}
\usepackage{tabularray}
\usepackage[normalem]{ulem}
\usepackage{enumitem}
\usepackage{pythonhighlight}
\usepackage[most]{tcolorbox}
\newtcolorbox{myquote}{colback=gray!10, colframe=gray!80, 
  left=0.6em, right=0.6em, top=0.3em, bottom=0.3em}

\newcommand{\ie}{\textit{i}.\textit{e}.}
\newcommand{\eg}{\textit{e}.\textit{g}.}
\newcommand{\etc}{\textit{etc}}
\newcommand{\update}[1]{\textcolor{blue}{#1}}
\newcommand{\revise}[1]{\textcolor{red}{#1}}

\begin{document}
\begin{sloppypar}

\title{AgentDP: Towards Automated Model Deployment for Heterogeneous Embedded AI Applications}

\author{Leming Shen\textsuperscript{1,3},
    Qiang Yang\textsuperscript{2},
    Xinyu Huang\textsuperscript{1},
    Zijing Ma\textsuperscript{1},
    Yuanqing Zheng\textsuperscript{1}
}

\affiliation{
    \textsuperscript{1}The Hong Kong Polytechnic University,
    \textsuperscript{2}U\textsuperscript{2}University of Cambridge,
    \textsuperscript{3}University College London
    \country{}
}

\email{
    {leming.shen, unixy-xinyu.huang, zijing.ma}@connect.polyu.hk, qy258@cam.ac.uk, csyqzheng@comp.polyu.edu.hk,
}

\renewcommand{\shortauthors}{Leming Shen, Qiang Yang, Xinyu Huang, Zijing Ma, Yuanqing Zheng}

\begin{abstract}
This paper presents \textit{AgentDP}, which aims to deploy pre-trained AI models on heterogeneous embedded devices by fully automating model optimization and adaptation (O\&A) with Large Language Model (LLM) agents. \textit{AgentDP} not only significantly reduces human effort via automation but also enhances model inference efficacy by integrating LLMs with device-specific domain knowledge. 
To deliver a robust automation framework, \textit{AgentDP} develops novel solutions to address three key technical challenges: 1) To help users elicit clear requirements, we propose a \textit{user requirement standardization} module; 2) To mitigate the execution environment mismatch between AI models and heterogeneous embedded devices, we design an \textit{operator-aware model partition} paradigm that enables different parts of the model to be efficiently executed on distinct inference engines. 3) To avoid re-searching optimization hyperparameters for new tasks, we propose a \textit{configuration reuse} module that heuristically retrieves previous hyperparameter configurations for efficient reuse.
Extensive experiments demonstrate that by fully automating the entire O\&A process, \textit{AgentDP} reduces entire model adaptation time by up to 80\% compared with manual O\&A while achieving higher task accuracy (more than 13\% on average) than the baselines. Additionally, a comprehensive user study also exhibits the robust generalizability of \textit{AgentDP} in adapting to different user requirements for heterogeneous embedded devices.

\end{abstract}



\acmYear{2025}\copyrightyear{2025}
\acmConference[Conference acronym 'XX]{Make sure to enter the correct
  conference title from your rights confirmation email}{June 03--05,
  2025}{Woodstock, NY}
\acmDOI{XXXXXXX.XXXXXXX}
\acmISBN{978-1-4503-XXXX-X/2018/06}

\maketitle
\section{Introduction}
\label{sec:introduction}


Embedded AI has emerged as a transformative paradigm, enabling the deployment of AI models directly on embedded devices such as Jetson Nano. Embedded AI facilitates real-time onboard data processing of various sensors (\eg, camera, microphone) and enables low-latency decision-making \cite{shen2025hierarchical, shen2026seeing}. In addition, embedded AI ensures privacy and security by keeping sensitive data on local devices \cite{shen2024fedconv}. 

\begin{figure}[t]
    \centering
    \includegraphics[width=0.48\textwidth]{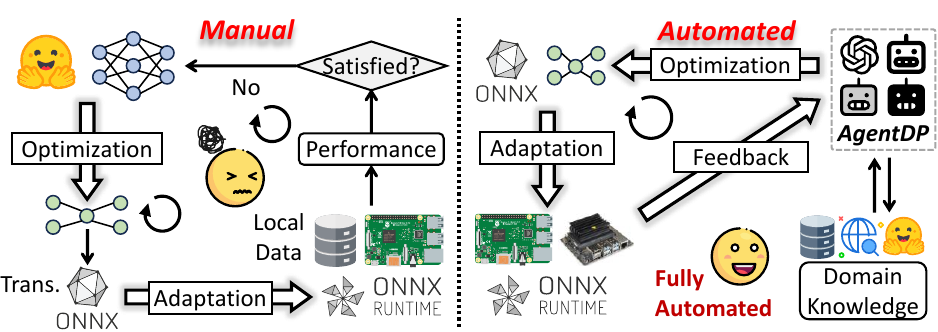}
    \caption{Deploying pre-trained open-source models on embedded devices (manual vs. automated).}
    \label{fig:manual_vs_agent}
    \vspace{-8pt}
\end{figure}



To develop an embedded AI application, developers can download a pre-trained model from open-source platforms (\eg, Hugging Face) and deploy the model on target embedded devices. These models, extensively trained on high-end servers using large-scale datasets (\eg, ImageNet \cite{deng2009imagenet}), can achieve high performance in various sensing tasks. In practice, however, most open-source models cannot be directly deployed on embedded devices for two main reasons. First, these models often contain a large number of parameters\footnote{Only a few popular models have downsized versions (\eg, YOLO-v5 \cite{redmon2016you}).}, making them hard to fit into resource-constrained embedded devices \cite{shen2024fedconv}. Second, these models are primarily available in PyTorch \cite{paszke2019pytorch} format, which may not be supported by embedded devices with heterogeneous execution environments and diverse dependencies. 
To tackle these practical challenges, developers need to perform multiple iterations of model optimization and adaptation, as illustrated in Fig.~\ref{fig:manual_vs_agent} (left).
To tailor models for resource-constrained devices, developers adopt various optimization techniques with tunable hyperparameters (\eg, pruning rate, quantization level, \etc.) to compress the models. To facilitate the portability of the compressed models across heterogeneous environments, developers also need to translate the compressed models into platform-independent \textit{portable models}, which can be deployed and executed on embedded devices via cross-platform inference engines (\eg, ONNX Runtime \cite{onnxruntime}).
Finally, developers can iteratively adapt the portable models to meet runtime performance requirements given the resource constraints of embedded devices\footnote{Although container techniques (e.g., Docker \cite{merkel2014docker}) can package dependencies to ease software portability, they often fail to transfer containers that pack customized AI models among embedded devices with heterogeneous CPU architectures \cite{docker_container_architecture}}.

Nevertheless, such an iterative O\&A process is cumbersome, error-prone, and inefficient.
1) \textbf{Large Search Space}: Manually defining and adjusting the search space for numerous tunable hyperparameters is time-consuming even when assisted by automated search frameworks \cite{yang2020hyperparameter}. For example, model pruning can be implemented with either channel-level or filter-level pruning with varying pruning ratios \cite{li2021hermes}. Worse still, highly specialized expertise is required to identify an appropriate search space for each hyperparameter. 
2) \textbf{Heterogeneous Target Devices}: Heterogeneous embedded devices have diverse execution environments, making the selection and installation of an appropriate inference engine with various dependencies challenging. For instance, TensorRT \cite{tensorrt} is optimized for Jetson Nano, whereas ONNX Runtime suits Raspberry Pis \cite{ahn2023performance} better. Mastering and manually configuring these device-specific frameworks demands substantial time and domain expertise \cite{pochelu2022deep}.
3) \textbf{Laborious Iterative Refinement}: Optimizing an AI model for high performance given the resource constraints of a target device typically necessitates multiple rounds of trial-and-error iterations, which is time-consuming even for experienced experts let alone novice developers. This motivates the following research question: \textit{\textbf{Can we automate the O\&A process and democratize the AI models that are readily open-sourced yet inaccessible for embedded AI applications?}} 


To fill this gap, our key insight is to employ LLM agents to fully automate the entire O\&A process. The recent advancement of LLMs (\eg, GPT-4 \cite{achiam2023gpt}) has demonstrated exceptional language comprehension and task automation capabilities. Embedded with extensive world knowledge, LLMs can effectively accomplish complex tasks (\eg, 
program synthesis \cite{shen2025gpiot, shen2025autoiot, shen2024iotcoder},  IoT data interpretation \cite{xu2024penetrative}, \etc.) in response to user requirements. Furthermore, by integrating LLMs with various third-party tools (\eg, web search tools, knowledge databases), we can empower LLM agents to facilitate task automation via enriched domain knowledge. To this end, we propose \textit{AgentDP}, a multi-agent-assisted model O\&A system for heterogeneous embedded AI applications. As shown in Fig.~\ref{fig:manual_vs_agent} (right), \textit{AgentDP} leverages multiple collaborative agents to automatically execute the iterative O\&A process.



Nevertheless, several technical challenges need to be addressed to deliver a functional and practical system: 1) \textbf{Vague User Requirements.} Due to complex and numerous model evaluation metrics, developers typically struggle to clearly and comprehensively articulate their requirements in a single attempt. Moreover, these requirements are often expressed in natural language with irregular structures, leading to misinterpretation of user requirements and real intents. 
To tackle this challenge, we propose a \textit{user requirement standardization} (URS) module that guides users to elicit their requirements following the Tree-of-Thought \cite{yao2023tree} approach tailored for the URS module. 
2) \textbf{Model-Engine Mismatch.} Though portable models are platform-independent, they may not be fully compatible with a single inference engine due to mismatched execution environments. For example, some AI models may use the \verb|Mod| operator during inference, which is not supported by the TensorRT engine \cite{tensorrt_mod_operator}. Developers have to either wait for engine updates or use an inefficient fallback solution (\eg, CPU).
To address this challenge, we propose an \textit{operator-aware model partition} module that allows different parts of a model to be executed using distinct engines. In addition, we design a \textit{cross-engine inference pipeline} to schedule the partitioned model for efficient inference.
3) \textbf{Exhausted Hyperparameter Search.} If \textit{AgentDP} performed model O\&A from scratch for all new tasks, the O\&A process would become extremely time-consuming due to the large hyperparameter search spaces. To mitigate this problem, we reuse hyperparameter configurations of previous models with similar resource profiles. Specifically, we model the hyperparameter tuning process as a finite state machine (FSM) and construct a graph database for similarity-based hyperparameter configuration retrieval.

We fully implement and evaluate \textit{AgentDP} on three heterogeneous embedded devices across various embedded AI applications, including image recognition, speech sentiment analysis, and vision-language understanding. We compare \textit{AgentDP} with two types of baselines: a SOTA LLM agent tailored for programming tasks and human developers. We invite 15 users as human baselines and evenly divide them into three groups based on their development experience and expertise. 
Extensive experiments demonstrate that \textit{AgentDP} can reduce the time required for the O\&A process by up to 61\% compared to manual execution by experts, significantly improving the model O\&A efficiency and reducing the human effort. Moreover, the portable models optimized by \textit{AgentDP} outperform baselines in terms of both task accuracy ($>$ 13\% on average) and resource consumption ($<$ 14\% on average). \textbf{Last but not least, by allowing users to freely express their requirements, our user studies demonstrate high generalizability (users rated overall satisfaction at over 4.8 out of 5) of \textit{AgentDP} in handling distinct requirements for adapting various open-source models to heterogeneous embedded devices.}

In summary, we make the following contributions:
\begin{itemize}[leftmargin=8pt]
    \item To our best, \textit{AgentDP} is the first system to fully automate the model O\&A process (from open-source model optimization to embedded AI application deployment), offering an end-to-end solution via multi-collaborative LLM agents.
    \item \textit{AgentDP} develops three novel technical modules for enhanced efficiency and robustness: a \textit{user requirement standardization} module for iterative user requirement elicitation, an \textit{operator-aware model partition} module with a \textit{cross-engine inference pipeline} to mitigate mismatch between models and inference engines, and a \textit{configuration reuse} module for efficient model optimization.
    \item Comprehensive evaluation validates the effectiveness and efficiency of \textit{AgentDP}, demonstrating that \textit{AgentDP} can significantly reduce the time required and enhance the performance for optimizing and adapting open-source models for embedded devices.
\end{itemize}

\section{Background \& Motivation}

\subsection{Challenges in Embedded AI}
\label{sec:2.1}
In this section, we provide a detailed analysis of why executing the iterative optimization and adaptation (O\&A) process for embedded AI is cumbersome, inefficient, and error-prone.

\noindent\textbf{Model Optimization.}
To reduce resource consumption, developers apply various optimization methods, each with distinct hyperparameter configurations. 
However, identifying an optimal hyperparameter set for model optimization is time-consuming even for experienced developers \cite{abraham2023homopt} due to two main reasons: 1) Determining appropriate hyperparameters from vast search spaces requires highly specialized expertise, as each hyperparameter influences the optimization result differently; 2) Multiple hyperparameter combinations may all satisfy the resource constraints of the target device, making the selection process non-trivial. For example, if the target device operates on battery power, hyperparameters that can minimize energy consumption may be prioritized.

\begin{figure}[t]
\setlength{\abovecaptionskip}{6pt}
    \centering
    \includegraphics[width=0.48\textwidth]{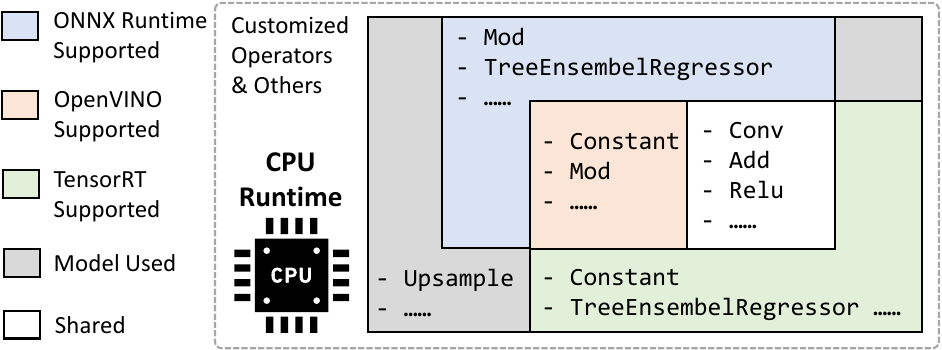}
    \caption{The mismatch between models and engines.}
    \label{fig:operator_set}
\vspace{-8pt}
\end{figure}

\noindent\textbf{Model Adaptation.}
To facilitate model portability, developers translate optimized models into portable models, which can be deployed on the embedded device via a cross-platform inference engine. Nonetheless, manually configuring model execution environments poses three key challenges: 1) Empirically following traditional configuration practices (\eg, directly deploying a PyTorch model on a server installed with commonly used environments) often fails due to hardware incompatibilities, dependency conflicts, and obsolete packages for embedded devices. For instance, some Jetson Nano boards may only support earlier and tailored versions of Ubuntu and Python environments \cite{jetson_nano_configuration}. 2) Embedded devices often exhibit varying affinities for different inference engines. For example, TensorRT is optimized for inference latency on Jetson Nano, whereas ONNX Runtime may be more suitable for Raspberry Pis. Mastering such highly specialized and device-specific domain knowledge is time-consuming even for experienced embedded AI developers. 3) 
Different models may employ a diverse set of operators (\eg, \verb|Add|, \verb|Conv|, \etc.) to perform various computation tasks. Similarly, different inference engines support distinct sets of operators based on their diverse architectures. 
In Fig.~\ref{fig:operator_set}, we see that none of the representative inference engines can fully support all operators of a Mask RCNN model \cite{he2017mask} as an example. 
This indicates that it is impractical to rely on a single inference engine for various embedded AI models on target devices. Moreover, it implies that models are typically executed in the less efficient fallback mode (\ie, CPU) in practice.


\noindent\textbf{Multiple Iterations.} 
The O\&A process typically undergoes multiple trial-and-error iterations by experts to optimize the model performance to meet user requirements. 
\textit{Such a labor-intensive process underscores the need for an automated system to enhance efficiency and reduce human effort.}


\subsection{LLM Agents}
The recent advancement in Large Language Models (LLMs) is fundamentally changing the application development process and task automation. 
By augmenting LLMs with diverse third-party tools (\eg, web search engines and knowledge databases), we can build intelligent agents \cite{shen2026one, li2026autoran, huang2026jailbreaking} to automate complex workflows with enriched external knowledge.

\begin{figure}[t]
\setlength{\abovecaptionskip}{6pt}
    \centering
    \includegraphics[width=0.48\textwidth]{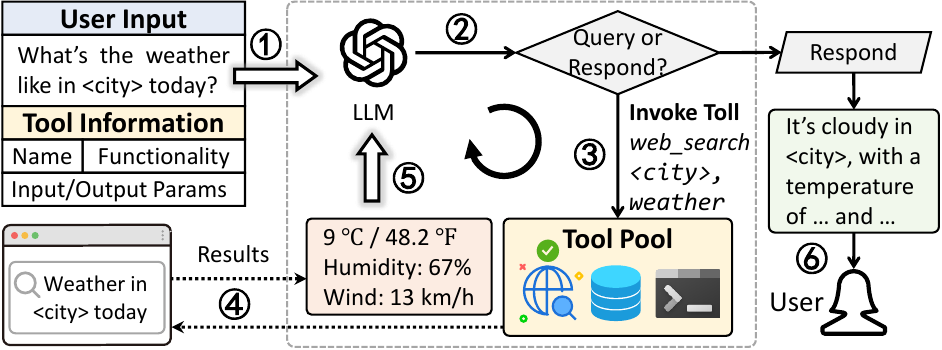}
    \caption{An illustrative workflow of LLM agents.}
    \label{fig:rationale_llm_agent}
    \vspace{-8pt}
\end{figure}

Fig.~\ref{fig:rationale_llm_agent} illustrates how an LLM agent can use various tools and respond to the user query, \eg, \textit{"What's the weather like in <city\_name>?"} Upon receiving the query, the agent encapsulates it with detailed metadata about the available tools, including their names, functionality descriptions, and input/output parameters (\ding{172}). For instance, a "web search" tool accepts keywords as input and outputs relevant search results. 
Based on this encapsulated prompt, the LLM agent decides whether to respond directly to the user or invoke any tools to retrieve additional information (\ding{173}). Since the user query involves \textit{real-time} weather, the LLM agent needs to invoke the tool with \verb|{"web_search": ["<city_name>", "weather"]}|
(\ding{174}). After receiving these parameters, the web search tool retrieves the requested weather (\eg, provided by National Weather Service) (\ding{175}) and returns the real-time result to the LLM agent (\ding{176}). Finally, the LLM agent parses the retrieved data and responds to the user (\ding{177}). Such a "query $\rightarrow$ tool call $\rightarrow$ respond" process can be fully automated. 

Furthermore, multiple agents can work together to solve complex tasks. 
For instance, SoA \cite{ishibashi2024self} enables automated large-scale code generation and optimization. As in real-world application development, complex programming tasks can be decomposed into sub-tasks and solved collaboratively.

\subsection{Agent Meets O\&A}
As discussed in \S~\ref{sec:2.1}, manually executing the O\&A process demands highly specialized domain expertise and substantial human effort, even for experienced developers. Therefore, we propose \textit{AgentDP} that fully automates the O\&A process by orchestrating multiple tool-enabled agents, offering unique advantages over manual model optimization and adaptation. 1) \textbf{Efficiency Enhancement.} \textit{AgentDP} achieves enhanced development efficiency, as it presents end-to-end full automation (from hyperparameter tuning and execution environment configuration to model deployment) with minimal human intervention.
2) \textbf{Reliability Enhancement.} \textit{AgentDP} improves the reliability of O\&A automation. 
Armed with information retrieval tools, \textit{AgentDP} can complement agents with domain knowledge needed in the O\&A process. \textit{AgentDP} also helps users elicit their requirements to avoid misunderstanding user intents.

\section{System Overview}

\begin{figure*}[t]
    \centering
    \includegraphics[width=\textwidth]{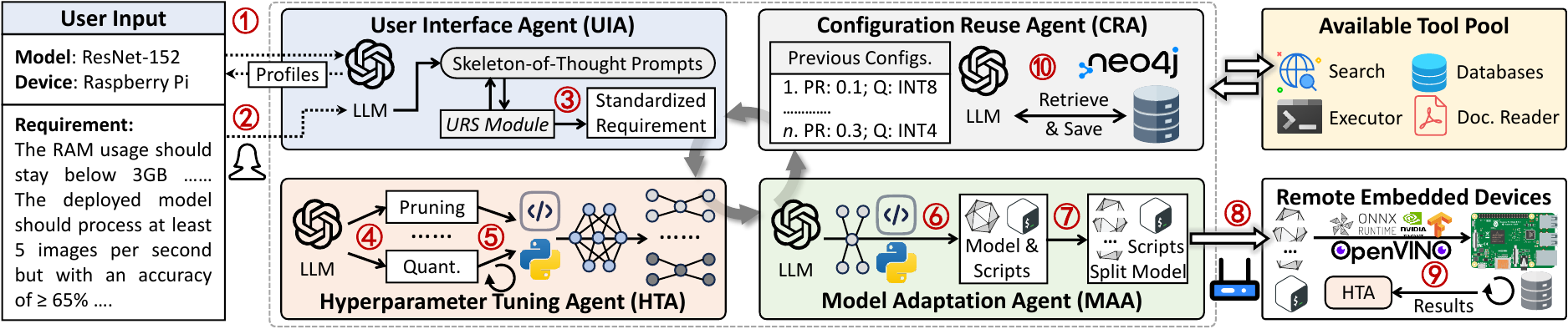}
    \caption{The system overview and workflow of \textit{AgentDP} with four specialized agents.}
    \label{fig:system_overview}
    \vspace{-8pt}
\end{figure*}

Fig.~\ref{fig:system_overview} illustrates the overall workflow of \textit{AgentDP}, consisting of four specialized agents 
that collaboratively execute the model O\&A process to fulfill the user requirements for embedded AI model deployment.

\noindent\textbf{Prerequisites.} 
1) \textit{AgentDP} uses cloud LLMs (\eg, GPT-4) for enhanced performance and robustness.
2) \textit{AgentDP} is deployed on an edge server and accesses the target device via a local network to enable remote access and control. 
3) Model inference and evaluation are entirely performed on target devices to preserve data privacy. Only inference results are transmitted to \textit{AgentDP} for further model optimization.
4) We provide the following tools for \textit{AgentDP}: a web search engine, a vector database, a graph database, a Python executor within a sandbox, and a document loader.


\noindent\textbf{Workflow.} 
From the user's perspective, she only needs to provide the model name, the target device name, and real-time performance requirements for deployment. Upon receiving this input, \textit{AgentDP} coordinates the four agents to automatically execute the O\&A process and present the final deployment results to the user.
\begin{itemize} [leftmargin=8pt]
    \item \textbf{User Interface Agent (UIA)} first constructs detailed profiles of the model and the target device (\ding{172}), serving as references for the user to elicit her requirements for deployment (\ding{173}). UIA then utilizes the \textit{user requirement standardization} module to iteratively refine and structure the requirement into a standardized format with precise semantics (\ding{174}).
    \item \textbf{Hyperparameter Tuning Agent (HTA)} is then guided to first generate a list of optimizable hyperparameters (\ding{175}) and then synthesize a script for automated hyperparameter tuning (\ding{176}). After iterative optimization, multiple optimized models that meet resource constraints are generated.
    \item \textbf{Model Adaptation Agent (MAA)} translates these models into portable models and generates a script to split the models into multiple parts via our \textit{operator-aware model partition} module (\ding{177}). MAA further synthesizes a script that can be executed on the target device to perform inference using our \textit{cross-engine inference pipeline} (\ding{178}). The partitioned model and the script are then transmitted to the target device (\ding{179}), which monitors the real-time performance using its local data and sends the results back to HTA as feedback to start a new optimization round (\ding{180}).
    \item \textbf{Configuration Reuse Agent (CRA)} further optimizes HTA by reusing and storing previous hyperparameter configurations (\ding{181}) during the model optimization process, thereby enhancing \textit{AgentDP}'s overall efficiency without searching from scratch.
\end{itemize}

\begin{figure}[t]
    \centering
    \includegraphics[width=0.45\textwidth]{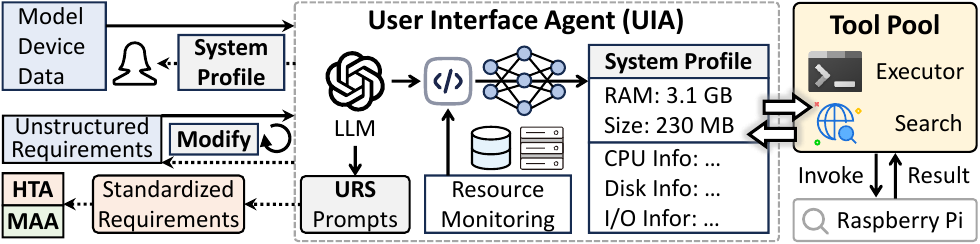}
    \caption{The workflow of User Interface Agent.}
    \label{fig:UIA_workflow}
    \vspace{-8pt}
\end{figure}

\noindent\textbf{Remarks.} 
The four agents closely collaborate by proactively invoking tools and sharing: model and device profiles, standardized user requirements, hyperparameter configurations, and model evaluation results on the embedded device.

\section{System Design}



\subsection{User Interface Agent (UIA)}
The heterogeneity of AI models and embedded devices makes it difficult even for experienced developers to clearly and comprehensively articulate their requirements using natural language (\eg, "The model should achieve an acceptable accuracy on the device"). This poses significant challenges for LLM agents to accurately interpret the requirements and verify whether they are ultimately fulfilled (\S~\ref{sec:URS_ablation}). To tackle this problem, we employ UIA to serve as the user interface to help users iteratively elicit clear and structured requirements.
As shown in Fig.~\ref{fig:UIA_workflow}, UIA contains two main modules, \ie, \textit{system profiling} and \textit{user requirement standardization} (URS).

\subsubsection{\textbf{System Profiling}}
\label{sec:system_profiling}
To provide users with a concise overview of the specified model and the target device, this module presents the resource consumption information of the model and the hardware capabilities of the target device.

\noindent\textbf{Model Profiling.}
Given the name of a large model as user input, UIA first invokes the web search tool to download its parameters from open-source platforms (\eg, Hugging Face). 
UIA then generates a script to deploy the model on a powerful server and perform inference tasks. The resource consumption information (\ie, model size, GPU memory usage, and RAM usage) is recorded and formatted as the model profile.

\noindent\textbf{Device Profiling.}
Given the name of a target embedded device (\eg, Jetson Nano) for model deployment, UIA invokes the web search tool to retrieve detailed specs of the device from its official website. Then, we utilize few-shot prompts \cite{brown2020language} to instruct UIA to process and format the retrieved contents into a structured device profile, including key specifications such as CPU, GPU, RAM, disk storage, connectivity, I/O ports, and operating system.

The constructed system profiles are further processed by the next \textit{user requirement standardization} module.

\subsubsection{\textbf{User Requirement Standardization (URS)}}
\label{sec:user_requirement_standardization}
Although using JSON or fill-in-the-blank UIs to capture clear user requirements appears promising, these methods often fail to accommodate the fluid and non-deterministic nature of user requirements. Instead, we adopt a Tree-of-Thought (ToT) \cite{yao2023tree} method to elicit clear and consistent requirements from the users through multiple rounds of UIA-user interaction. Intuitively, the ToT-based approach helps users sort out ideas and facilitates systematic refinement of user requirements through multi-path reasoning, effectively resolving ambiguities and inconsistencies. Our design includes two main considerations: 1) model deployment often involves multiple performance metrics and even experienced users may not clearly express the requirements within a single attempt; 2) users often express vague or inconsistent requirements that can lead to suboptimal model selection or system failures.

\begin{figure}
    \centering
    \subfigure[Prompt for ambiguity detection]{
        \includegraphics[width=0.225\textwidth]{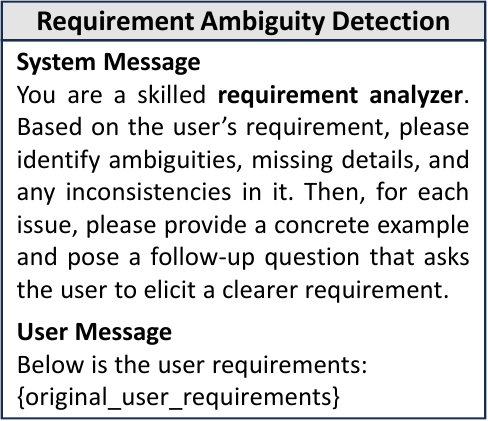}
        \label{fig:ambiguity_detection_prompt}
    }
    \centering
    \subfigure[Standardized user requirement]{
        \includegraphics[width=0.225\textwidth]{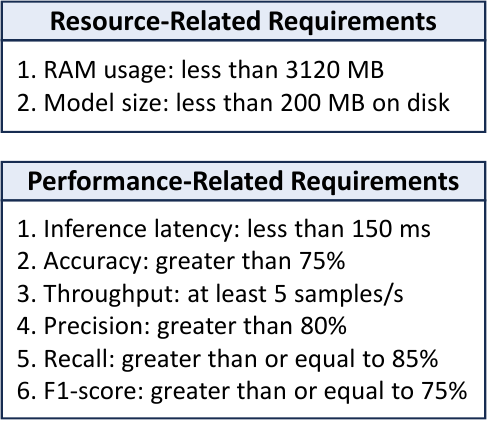}
        \label{fig:standardized_user_requirement}
    }
    \caption{(a) \textit{AgentDP} prompts UIA to identify ambiguities and request the user to make clarification; (b) The standardized user requirements with clear structures.}
    \vspace{-8pt}
\end{figure}

We first categorize user requirements for model deployment into two aspects:
(1) \textit{Resource requirements} are task-agnostic that specify resource constraints for model deployment and inference, including RAM usage, GPU memory usage, and model size (\ie, the storage size of model parameters on local disks). (2) \textit{Performance requirements} specify various metrics (\eg, accuracy, precision, throughput, \etc.) evaluated during real-time inference on the target device. 
Next, to help the user iteratively elicit her requirements progressively, UIA leverages a ToT-based approach to exploit multiple reasoning paths simultaneously, enabling more structured, coherent, and accurate decision-making. \textbf{ToT 1}: UIA presents the obtained system profiles for the user as references. \textbf{ToT 2}: UIA analyzes the system profiles and generates a list of possible metrics for the user. For example, the metrics may include classification accuracy for image recognition models and tokens per second for language models. The user can then selectively specify her requirement for those metrics. \textbf{ToT 3}: \textit{AgentDP} prompts (Fig.~\ref{fig:ambiguity_detection_prompt}) UIA to detect ambiguities, missing details, and inconsistencies in the requirements (\eg, "\textit{The model should achieve a high accuracy}"). Based on the detection results, UIA is further instructed to provide examples of well-defined requirements and pose follow-up questions to elicit further clarification from the user. For instance, "\textit{Example: The model should achieve a classification of at least 90\%; Question: What is the minimum proportion of successfully recognized images?}" 

Moreover, UIA is explicitly prompted to identify any requirements that conflict with the system profile and accordingly suggest appropriate values for the user. For instance, if the original model achieves a maximum throughput of five samples per second on the edge server, the expected value on the target embedded device should not exceed this limit. Through multiple iterations of the "requirement $\rightarrow$ detection $\rightarrow$ question $\rightarrow$ clarification" cycle, user requirements are progressively refined into a clear format. As shown in Fig.~\ref{fig:standardized_user_requirement}, each line consists of two elements: an evaluation metric and an expected value. The standardized user requirements are then passed to HTA and MAA for further processing.

\vspace{4pt}
\noindent\textbf{Remarks.} By converting vaguely expressed user requirements into well-structured formats with clear semantics, \textit{other agents can effectively parse and verify them during iterative optimization and evaluation of the deployed model.}


\subsection{Hyperparameter Tuning Agent (HTA)}
Given the system profiles and the standardized user requirements from UIA, HTA tailors models for resource-constrained devices by tuning various hyperparameters for optimization. 
However, determining appropriate search spaces for various hyperparameters and selecting optimal sets requires highly specialized domain expertise (\S~\ref{sec:2.1}). To tackle this issue, we meticulously design several ToT prompts and explicitly instruct HTA to comprehensively analyze the system profiles and resource requirements.  
As shown in Fig.~\ref{fig:HTA_workflow}, HTA contains the following three stages: 

\begin{figure}[t]
\vspace{4pt}
    \centering
    \includegraphics[width=0.45\textwidth]{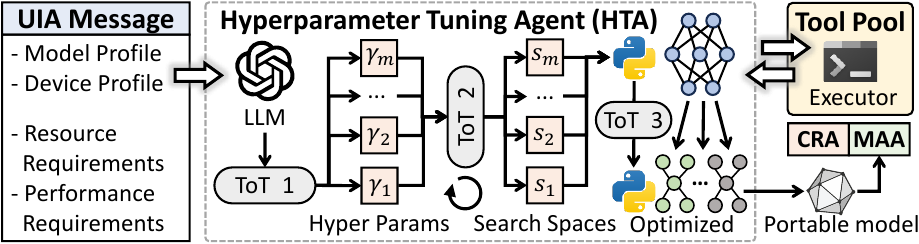}
    \caption{The workflow of Hyperparameter Tuning Agent.}
    \label{fig:HTA_workflow}
    \vspace{-8pt}
\end{figure}

\begin{figure}
    \centering
    \subfigure[The prompt template for tunable hyperparameter determination]{
        \includegraphics[width=0.225\textwidth]{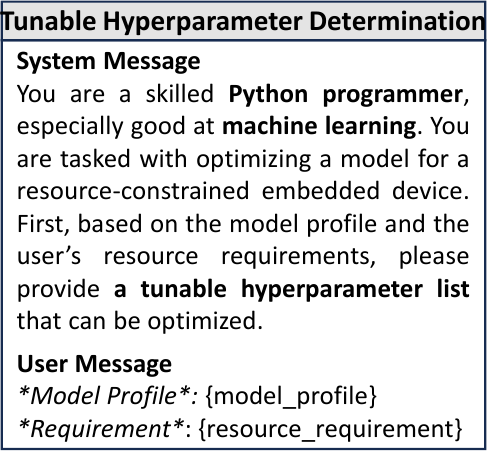}
        \label{fig:optimizable_hyperparameter_determination_prompt}
    }
    \centering
    \subfigure[The prompt template for search space generation]{
        \includegraphics[width=0.225\textwidth]{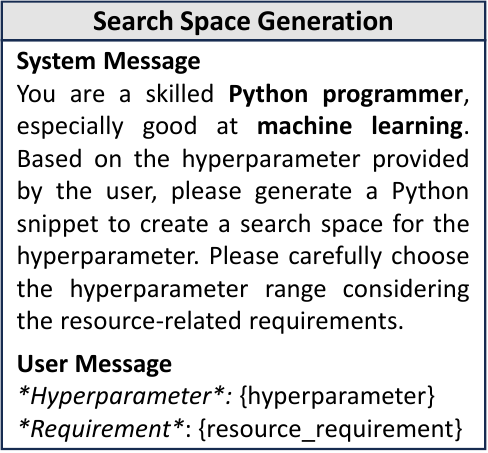}
        \label{fig:search_space_generation_prompt}
    }
    \caption{Example of ToT-based prompts.}
    \vspace{-8pt}
\end{figure}

\noindent{\textbf{Tunable Hyperparameter Determination (ToT 1)}}
\textit{AgentDP} prompts (Fig.~\ref{fig:optimizable_hyperparameter_determination_prompt}) HTA to first generate a comprehensive tunable hyperparameter set (\eg, batch size, pruning rate, quantization level, \etc.) that can reduce resource consumption of the model. This step can be regarded as identifying multiple branches (\ie, tunable hyperparameters) within the optimization tree. Next, HTA is instructed to filter out hyperparameters that modify model architectures (\eg, layer number, activation function). Adjusting these hyperparameters in a well-trained model may cause unexpected performance degradation, as it disrupts the learned feature representations, leading to a mismatch between the new architecture and previously optimized weights \cite{liu2018rethinking}.

\noindent{\textbf{Search Space Generation (ToT 2)}}
For each filtered hyperparameter, \textit{AgentDP} further employs Chain-of-Thought (CoT) \cite{wei2022chain} prompts that instruct HTA to: 1) Analyze the resource profiles of the original large model and the corresponding resource requirements; 2) Generate an appropriate search range for the hyperparameter (Fig.~\ref{fig:search_space_generation_prompt}). For instance, if the parameter size of the original model is 100 MB while the user specifies that the parameter size of the optimized model should be less than 80 MB, the pruning rate can range within $[0.2, 0.5]$. Moreover, HTA is instructed to relax the range a little (\eg, $[0, 0.5]$), considering that other optimization methods are simultaneously applied during optimization.

\noindent{\textbf{Optimal Hyperparameter Searching (ToT 3)}}
\label{sec:hyperparameter_searching}
\textit{AgentDP} prompts HTA to create a comprehensive multi-objective optimization function that takes tunable hyperparameters as input variables, aiming to minimize the resource consumption of the optimized model. Accordingly, HTA generates a script that performs iterative hyperparameter tuning to identify multiple hyperparameter sets that satisfy all resource requirements. Finally, HTA synthesizes a script that translates the original model into multiple optimized portable models by applying the identified hyperparameter sets. The portable models are then sent to MAA for model adaptation.

\vspace{4pt}
\noindent\textbf{Remarks.} 
Hyperparameter filtering and appropriate search space determination leverage LLMs' extensive world knowledge, enhancing the \textit{\textbf{reliability and efficiency}} of model optimization. Notably, more advanced hyperparameter searching algorithms (\eg, BOHB \cite{falkner2018bohb}) can also be further integrated orthogonally in the script synthesized by HTA for enhanced model optimization performance.

\begin{figure}
    \centering
    \includegraphics[width=0.45\textwidth]{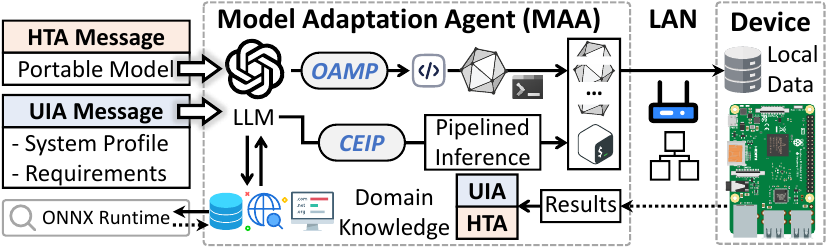}
    \caption{The workflow of Model Adaptation Agent.}
    \label{fig:MAA_workflow}
    \vspace{-8pt}
\end{figure}

\subsection{Model Adaptation Agent (MAA)}
To cater to heterogeneous embedded AI environments, we need to adapt portable models to the target device via various inference engines. However, the operators used by portable models may not be fully supported by a single inference engine (\S~\ref{sec:2.1}). 
Existing solutions either rely on fallback environments (\eg, CPU) for model execution or continuously develop new cross-platform frameworks to support additional operators, both of which are inefficient and cumbersome.
To tackle this problem, we propose MMA (Fig.~\ref{fig:MAA_workflow}), which features an \textit{operator-aware model partition} module that divides the portable model into multiple partitions, each compatible with a different inference engine. Accordingly, MMA also employs a \textit{cross-engine inference pipeline} to arrange the model partitions and perform inference tasks on the target device by switching between different engines.


\subsubsection{\textbf{Operator-Aware Model Partition (OAMP)}}
\begin{figure}
    \centering
    \subfigure[DAG \& Operators]{
        \includegraphics[width=0.225\textwidth]{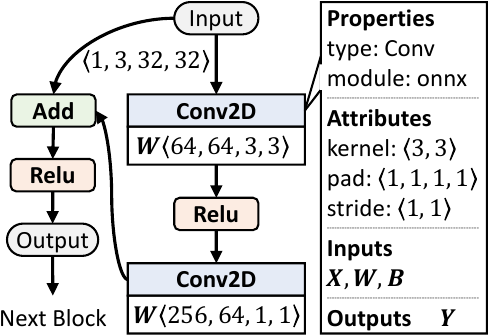}
        \label{fig:DAG}
    }
    \centering
    \subfigure[An example of CEIP]{
        \includegraphics[width=0.225\textwidth]{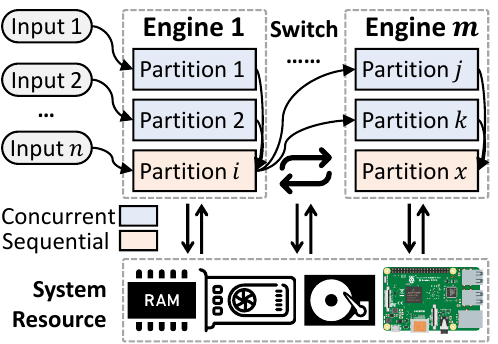}
        \label{fig:scheduling}
    }
    \caption{(a) A sample DAG with its operators from part of a portable model; (b) Our CEIP module.}
\vspace{-8pt}
\end{figure}

Generally, a portable model is represented as a Directed Acyclic Graph (DAG), where nodes correspond to operators (\eg, \verb|Add|, \verb|Conv|, \verb|Relu|), and edges represent tensors that store data flowing between these operators. Fig.~\ref{fig:DAG} provides a DAG example of a residual block from a portable model. Each operator in the DAG includes its properties (\eg, type), attributes (\eg, padding), and I/O parameters. Such a graph structure can be parsed by different inference engines to perform inference tasks according to the DAG. Nonetheless, the operator set supported by a single inference engine may not cover all the operators used in a portable model. For example, TensorRT does not support the \verb|Mod| operator while ONNX Runtime does \cite{tensorrt_mod_operator, onnxruntime_mod_operator}. To address this problem, we propose the OAMP module to split a portable model into multiple parts, ensuring that each partition can be executed on different but compatible inference engines. Specifically, our OAMP module contains two key stages: the \textit{operator availability check} and the \textit{greedy partitioning algorithm}.


The \textit{\textbf{operator availability check}} analyzes the operators used by the portable model and those supported by distinct inference engines. Specifically, \textit{AgentDP} first instructs MAA to synthesize a tailored script that analyzes the DAG of the portable model and produces a list of the operators used. Next, MAA is prompted to generate a list of inference engines compatible with the target device. For each engine, MAA invokes the web search tool to retrieve a detailed list of supported operators. In addition, MAA estimates the inference latency of each operator across all supported engines.


The \textit{\textbf{greedy partition algorithm}} aims to minimize the total inference latency while preserving operators' dependency consistency. Specifically, for each operator used by the portable model, three cases are considered: 1) If the operator is supported by multiple inference engines, it is assigned to the one with the lowest latency (\eg, TensorRT for Jetson Nano). 2) If the operator is supported by only one engine, it is directly assigned to that engine. 3) If the operator is not supported by any available engine, it is assigned to a fallback execution environment (\eg, CPU). Furthermore, considering the overhead incurred when transferring data and control signal between different engines, \ie, context switching \cite{yang2023edgefm}, we introduce an additional condition. When assigning an engine different from the one used for the previous operator, we calculate the sum of the context-switching time and the inference latency using the new engine. If the computed time exceeds the inference time using the previous engine, the current operator is assigned to the previous engine; otherwise, it is assigned to the new engine. After such a greedy approach, MAA generates a list of model partitions with their corresponding inference engines.
This method ensures that the model partitioning process is both efficient and adaptive to the heterogeneous capabilities of different inference engines in embedded environments while minimizing the performance penalties caused by context switching. 



\subsubsection{\textbf{Cross-Engine Inference Pipeline (CEIP)}}
To minimize context switching overhead and ensure efficient data flow between model partitions during inference, we propose the CEIP module to implement a pipelined inference paradigm across multiple engines, as illustrated in Fig.~\ref{fig:scheduling}. Specifically, MAA first loads each model partition onto its assigned inference engine, ensuring that memory and data alignment are consistent across different execution environments. Inference is then performed sequentially or concurrently, depending on the dependency structure of the portable model's DAG. Data transfer between partitions is optimized using shared memory and direct memory access (DMA) to minimize context switching time. Additionally, before context switching, CEIP pre-fetches the input of the next partition while the current partition is still executing. This design effectively exploits the strengths of different inference engines (\eg, TensorRT's GPU acceleration) to maximize inference efficiency and reduce execution latency.

\begin{figure}[t]
    \centering
    \includegraphics[width=0.48\textwidth]{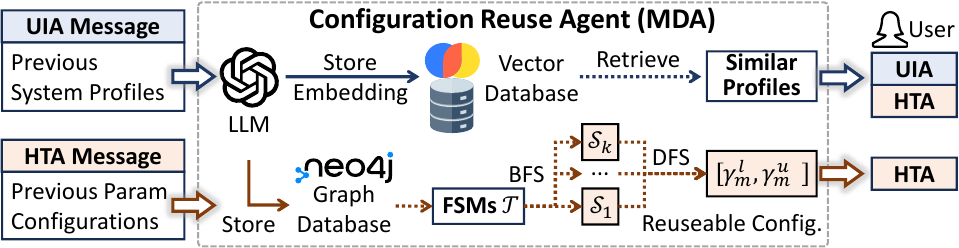}
    \caption{The workflow of configuration reuse agent.}
    \label{fig:CRA_workflow}
    \vspace{-8pt}
\end{figure}

\subsubsection{\textbf{Model Evaluation}}
With the model partitions and the inference pipeline, the target device performs inference on its local data and monitors the real-time performance, which is then sent to MAA. \textit{\textbf{For each portable model's monitoring results, MAA judges whether all performance requirements are satisfied. If they are met, the hyperparameter configuration is delivered to CRA and the monitoring results are presented to the user.}} Till now, a single round of the model O\&A process is completed. The user can either select a preferred configuration for model deployment or choose to continue optimizing the model, in which case \textit{AgentDP} will initiate a new round and repeat the automated O\&A process until the user is satisfied with the final deployment result.

\vspace{4pt}
\noindent\textbf{Remarks.} As the supported operators evolve with version updates and iterations of different inference engines, all the scripts and algorithms need to be automatically and dynamically implemented by MAA through our meticulously designed prompts to deal with heterogeneous embedded AI environments. The OAMP and CEIP modules fully exploit the extensive world knowledge and the program synthesis capabilities of LLM agents, enhancing the reliability and efficiency of model inference on embedded devices. Note that more advanced and efficient model inference strategies (\eg, prefetching \cite{jang2024eden} and virtual memory \cite{jeon2025tinymem}) can also be orthogonally integrated for enhanced efficiency.

\subsection{Configuration Reuse Agent (CRA)}
During model optimization, we observe that hyperparameter tuning often becomes time-consuming, though much faster than manual tuning. This is because \textit{AgentDP} currently generates a completely new search space for each user request, resulting in heavy computational overhead, long waiting times, and poor user experiences. \textit{Can we progressively accelerate hyperparameter tuning to further enhance the efficiency of \textit{AgentDP}?} Our insight is that, as system utilization increases, \textit{AgentDP} accumulates numerous historical hyperparameter configurations, which can potentially be reused for new but similar models. To this end, as shown in Fig.~\ref{fig:CRA_workflow}, we develop CRA that maintains two databases: one for \textit{system profiles} and another for \textit{hyperparameters}.

\subsubsection{\textbf{Profile Reuse}}
Model profiles record the resource consumption of the original large model, while device profiles include specs of the target device (\S~\ref{sec:system_profiling}).
For each unique model and embedded device, CRA stores the constructed profiles in a vector database, respectively. When a user specifies the same model or device, CRA retrieves the previously stored profiles from the database and delivers them to UIA and HTA for further processing.

\subsubsection{\textbf{Hyperparameter Reuse}}
We model each hyperparameter tuning process as a finite state machine (FSM) \cite{lee1996principles} and store the constructed FSM in a graph database. When optimizing a new large model, CRA first retrieves several similar FSMs and then accordingly determines narrowed search spaces for efficient hyperparameter tuning.

\begin{figure}
    \centering
    \includegraphics[width=0.47\textwidth]{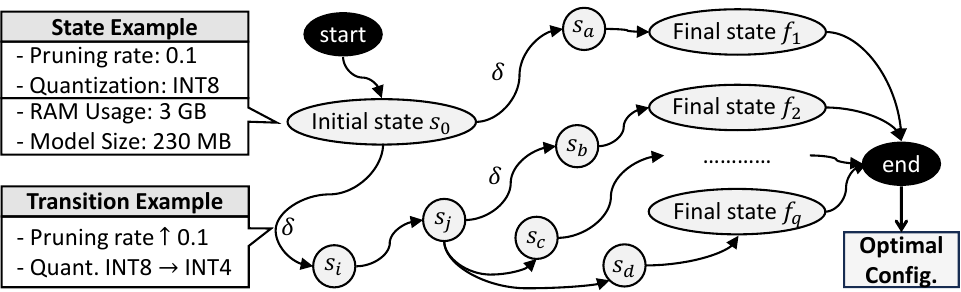}
    \caption{FSM for hyperparameter tuning modeling.}
    \label{fig:FSM}
    \vspace{-8pt}
\end{figure}

\noindent\textbf{FSM-Based Modeling.}
Fig.~\ref{fig:FSM} illustrates an example of an FSM, where each node represents a state (\ie, an optimized model), including a hyperparameter configuration and the corresponding resource consumption. The initial state corresponds to the original model and the terminal states represent a set of optimized models that satisfy all resource requirements. The directed line between two nodes denotes a state transition that occurs when a hyperparameter is modified during tuning. For example, when the pruning rate increases from 0.1 to 0.2, the memory usage of the optimized model may change from 90 MB to 80 MB. As such, an FSM can be constructed and regarded as a unique feature for each model after hyperparameter tuning, which is represented as:
\begin{equation}
\label{eq:FSM}
    \begin{array}{l}
        H=(S, s_0, \delta, F)\\
        s_i\stackrel{\delta}{\longrightarrow}s_j,\; \forall s_i\in S, \forall s_j\in S \\
        F = \{f_1, f_2, \cdots, f_q\} \subset S
    \end{array}    
\end{equation}
where $S$ is a set of states (\ie, optimized models), $s_0$ represents the initial state, $\delta$ is the state transition function that transitions from $s_i$ to $s_j$ by modifying tunable hyperparameters, and $F$ is a set of final states. With the automatically constructed FSM by HTA after hyperparameter tuning, CRA stores it into a dynamically maintained graph database $\mathcal{D}$.

\noindent\textbf{Similar FSM Retrieval.} 
Given the profile of a new model for optimization, CRA searches the FSM database and retrieves the top-$k$ FSMs whose initial resource consumption profiles are similar to the current one. Specifically, considering each resource metric (\eg, model size, RAM, GPU memory) in the profile, we define resource profile similarity (RPS) between the new model and each FSM in the database as:
\begin{equation}
\label{eq:RPS}
    RPS_i=\frac{|r_i^1 - r^1|}{r^1}+\frac{|r_i^2 - r^2|}{r^2}+\cdots+\frac{|r_i^N - r^N|}{r^N}\; \forall r_i^j\in \mathcal{D}
\end{equation}
where $r^N$ is the consumption of the $N$-th resource metric of the new model, $r_i^N$ is the consumption of the $N$-th resource metric of the $i$-th FSM. A smaller RPS value indicates a higher similarity between the new model and the FSM. The selected top-$k$ similar FSMs are denoted as $\mathcal{T}=\{H_1, H_2, \cdots, H_{k}\}$.

\noindent\textbf{Hyperparameter Reuse.}
Given a list of similar FSMs retrieved from the database, CRA will find a set of appropriate hyperparameters for reuse. Specifically, CRA first performs a breadth-first search (BFS) starting from the initial state of each selected FSM and obtains a list of states including their hyperparameter configurations:
\begin{equation}
\label{eq:BFS}
    \begin{array}{l}
        \mathcal{T}=\{\mathcal{S}_1, \mathcal{S}_2, \cdots, \mathcal{S}_{k}\},\;\forall \mathcal{S}_i\subset S_i\\
        \mathcal{S}_i = \{s=\{\gamma_1^i, \gamma_2^i, \cdots, \gamma_m^i\}\;|\;\forall s\; \text{adjacent to}\;s_0^i,\;s_0^i\in S_i\}\\
        \Gamma_m=\{\gamma_m^1, \gamma_m^2, \cdots, \gamma_m^i\}, \; \forall \gamma_m^i \in \mathcal{S}_i
    \end{array}
\end{equation}
where $\mathcal{S}_i$ is a set of states adjacent to the initial state $s_0^i$ from the $i$-th selected FSM, $\gamma_m^i$ is the value of the $m$-th hyperparameter of an adjacent state, and $\Gamma_m$ is a list of the $m$-th hyperparameter from all adjacent states. For each hyperparameter in $\Gamma_m$, the lower bound of the $m$-th hyperparameter's search space is defined as the minimum value of the list:
\begin{equation}
\label{eq:BFS}
    \gamma_m^l = \min(\Gamma_m)
\end{equation}
The rationale behind such design is that, during tuning, hyperparameters are typically initialized with small values to minimize initial performance degradation \cite{weng2021neural}. Consequently, the states obtained after BFS contain smaller values among all the hyperparameters.
Next, CRA conducts a depth-first search (DFS) to traverse the entire FSM and record the maximum value of the $m$-th hyperparameter, serving as the upper bound of its search space:
\begin{equation}
\label{eq:BFS}
    \gamma_m^u = \max\{\gamma_m^1, \gamma_m^2, \cdots, \gamma_m^j, \cdots\}
\end{equation}
where $\gamma_m^j$ denotes the $j$-th value of the $m$-th hyperparameter value within the traversed path. As such, the search space for the $m$-th hyperparameter is $[\gamma_m^l, \gamma_m^u]$, which is narrower compared to creating a new search space.

\vspace{4pt}
\noindent\textbf{Remarks.} We design meticulously crafted prompts and interfaces for CRA to automatically access the graph database and analyze the FSMs. Experiments (\S~\ref{sec:ablation_reuse}) reveal that CRA can reduce the time required for model optimization by up to 74\%, significantly enhancing both user experience and system efficiency with reduced computation overhead.

\section{Experiment Setup}

\subsection{Implementation}
\noindent\textbf{Hardware.}
We implement and deploy \textit{AgentDP} on an edge server and deploy various models on three embedded devices: a Raspberry Pi, a Jetson Nano, and an embedded mini PC. The edge server and the embedded devices are connected via a router within a local area network. Table~\ref{tb:edge_device_information} shows the hardware configurations of the devices.


\noindent\textbf{Software.}
We develop \textit{AgentDP} with 2,683 lines of Python code. We use LangChain \cite{Chase2022LangChain} to construct LLM agents. Tavily \cite{Asher2022tavily} and Firecrawl \cite{Camara2022firecrawl} are selected as the web search tools. We choose Chroma \cite{Bashir2022chroma} as the vector database for storing system profiles and use Neo4j \cite{Finne2022neo4j} as the graph database to store FSMs for efficient hyperparameter reuse. We implement a communication interface for \textit{AgentDP} to support remote access and control of target devices. We also build a Python executor within a sandbox \cite{docker}. GPT-4o \cite{hurst2024gpt} serves as the backend LLM of \textit{AgentDP} and text-embedding-ada-002 \cite{openai2022textembeddingada} is used for embedding. We integrate the Optuna package \cite{akiba2019optuna} with \textit{AgentDP} for HTA to find multiple optimal hyperparameter sets. For embedded AI models, we use PyTorch \cite{paszke2019pytorch} framework to run the original model and translate it into ONNX \cite{onnx} as the portable model. We implement our \textit{cross-engine inference pipeline} considering the following inference engines: ONNX Runtime \cite{onnxruntime}, TensorRT \cite{tensorrt}, OpenVINO \cite{openvino}, TVM \cite{chen2018tvm}, and MLIR \cite{jin2020compiling}. Reflexion \cite{shinn2023reflexion} is used to ensure the correctness of intermediate outputs.

\subsection{Embedded AI Applications \& Datasets}
\textit{Based on the input data modality, model architecture, and device heterogeneity, we select three representative embedded AI applications: image recognition (visual), speech sentiment analysis (audio), and visual-language understanding (multimodal).}
1) \textbf{Image Recognition (IR)} on embedded devices enable real-time processing and analysis of \textit{visual data} in areas such as autonomous driving \cite{he2023vi} and healthcare \cite{ouyang2023harmony}. We select a pre-trained \textit{CNN}-based ResNet-152 \cite{he2016deep} model and instruct \textit{AgentDP} to deploy it on a \textit{Raspberry Pi}. We evaluate the deployed model on a subset of the ImageNet dataset \cite{deng2009imagenet}, containing 100,000 RGB images in 1,000 classes. 2) \textbf{Speech Sentiment Analysis (SSA)} on \textit{audio data} involves analyzing speech patterns and pitch to determine the emotional tone of spoken content, enabling smart voice assistants, customer service interactions, \etc. We select the fine-tuned \textit{CNN+Transformer}-based Whisper model\footnote{\url{https://huggingface.co/firdhokk/speech-emotion-recognition-with-openai-whisper-large-v3}} and deploy it on a \textit{Jetson Nano}. We evaluate the deployed model using the TESS \cite{SP2/E8H2MF_2020} dataset that has 2,800 audio samples of seven types of emotions. 3) \textbf{Vision-Language Understanding (VLU)} leverages \textit{Vision-Language Models} (VLMs) to interpret and describe general images (\eg, logical diagrams and formula recognition), enabling applications such as visually impaired assistant \cite{zhang2025vlm}. We select the pre-trained multimodal \textit{Transformer}-based DeepSeek-VL-1.3b \cite{lu2024deepseek} as the VLM model and prompt \textit{AgentDP} to deploy it on an embedded mini PC with 12 GB of GPU memory. We evaluate the deployed model on the MMMU benchmark \cite{yue2024mmmu}, containing 11.5K meticulously collected image understanding questions covering six core disciplines, \ie, art \& design, business, science, medicine \& healthcare, humanities, and technology \& engineering. Note that the fine-tuned Whisper model and the VLM use certain operators (\eg, \verb|LayerNormalization|, \verb|FusedConv|) that are not fully supported by any single inference engine.

\section{Evaluation}

\begin{table}[t]
\small
\caption{Detailed information of the embedded devices.}
\label{tb:edge_device_information}
\begin{tabular}{ccccc}
\toprule[1pt]
Type                         & Device         & RAM   & VRAM & Storage \\ \toprule[1pt]
\multirow{3}{*}{\makecell{Embedded\\Device}} & Raspberry Pi 5 & 8GB   & -    & 32GB      \\
                             & Jetson Nano    & 4GB   & 4GB  & 64GB      \\
                             & IBOX-606 mini PC    & 32GB  & 12GB & 256GB     \\ \hline
Router                       & Mi AX3000      & 256MB & -    & -         \\ \hline
Edge Server                      & Supermicro X11SCA-F  & 32GB  & 16GB  & 1T        \\ \bottomrule[1pt]
\end{tabular}
\vspace{-4pt}
\end{table}

\begin{figure*}[t]
    \centering
    \subfigure[System profiles (simplified)]{
        \includegraphics[width=0.225\textwidth]{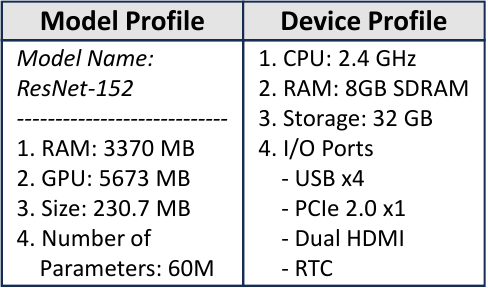}
        \label{fig:image_recognition_system_profile}
    }
    \subfigure[Resource consumptions]{
        \includegraphics[width=0.225\textwidth]{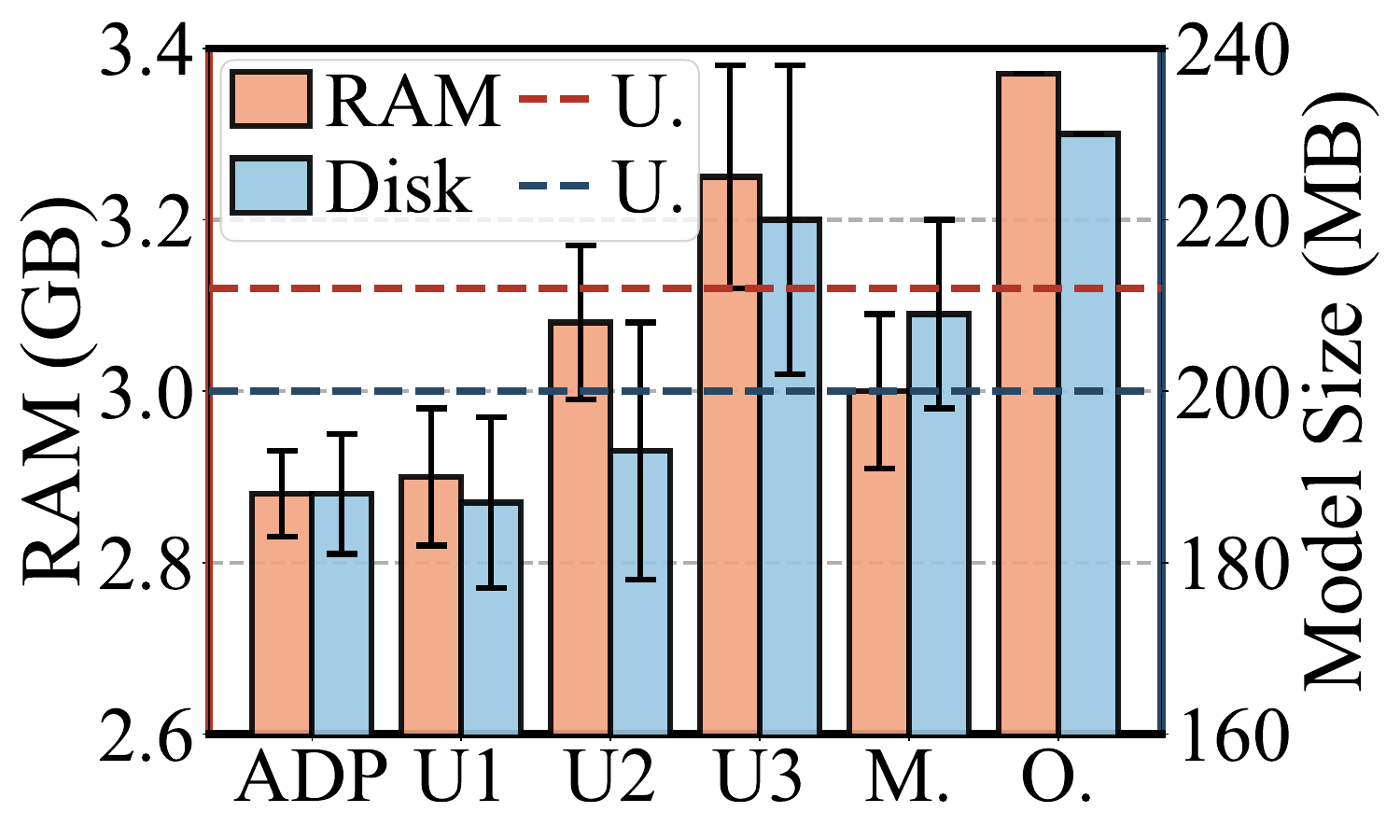}
        \label{fig:image_recognition_resource_consumption}
    }
    \centering
    \subfigure[Task accuracy]{
        \includegraphics[width=0.225\textwidth]{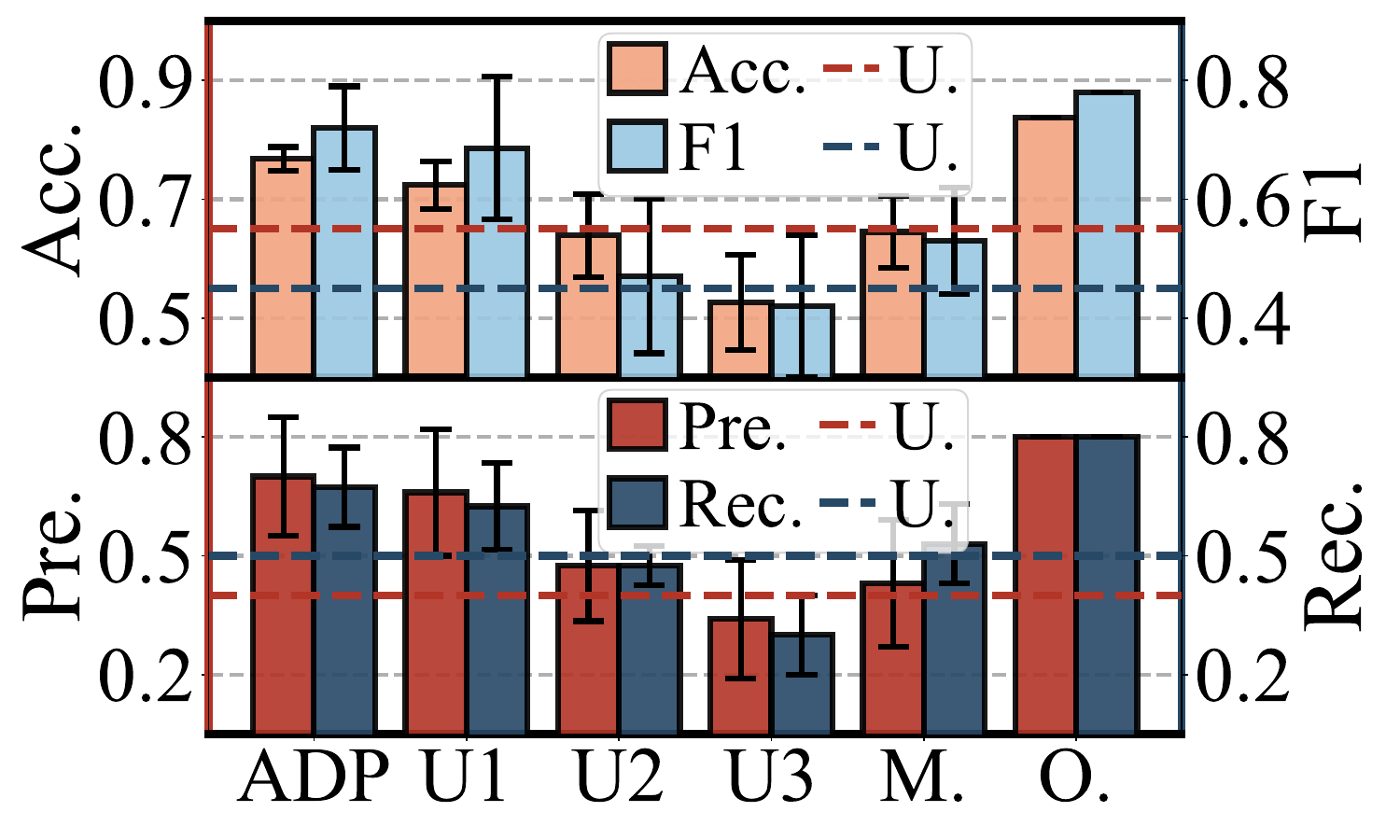}
        \label{fig:image_recognition_task_accuracy}
    }
    \subfigure[Real-time performance]{
        \includegraphics[width=0.225\textwidth]{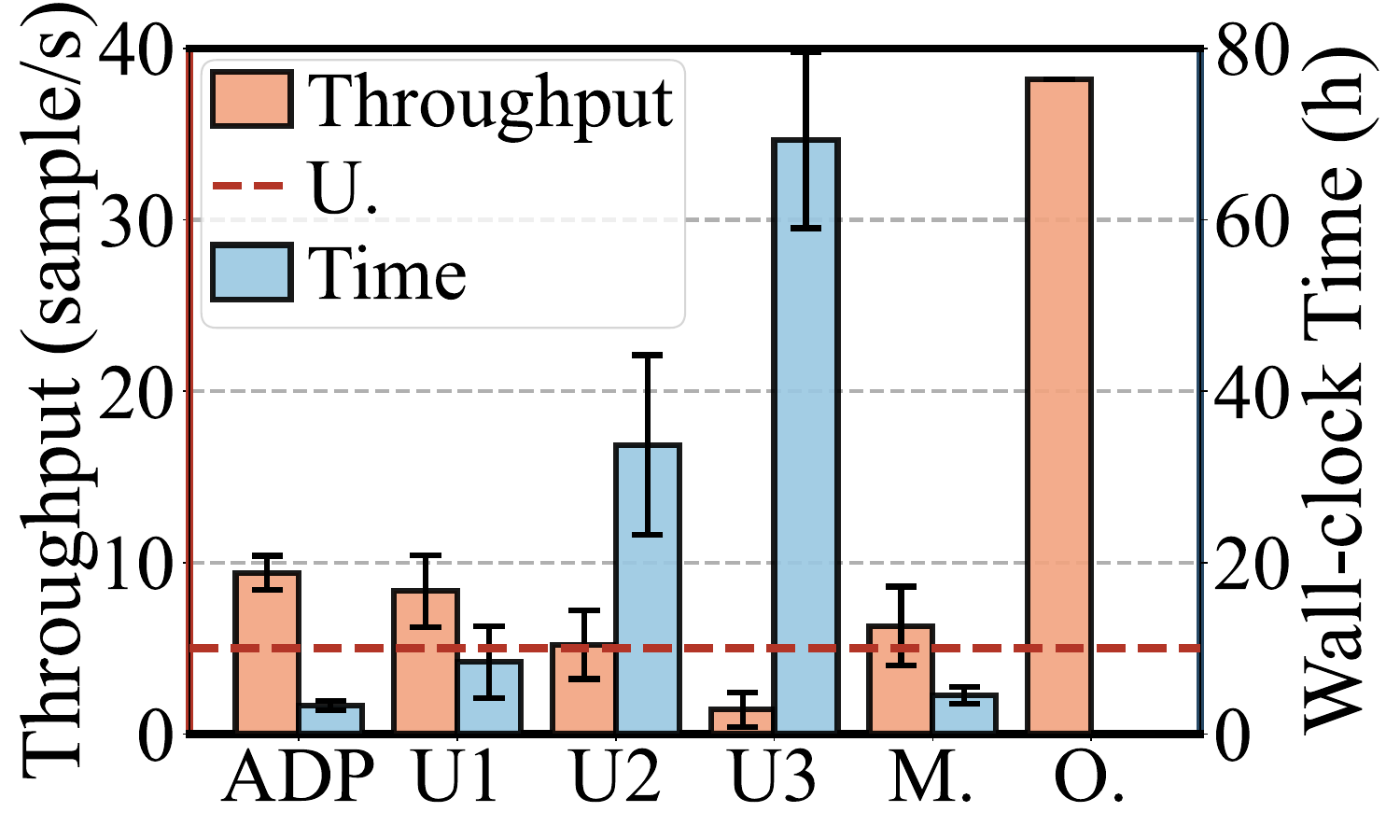}
        \label{fig:image_recognition_realtime_performance}
    }
    \caption{Evaluation results of image recognition (ADP for \textit{AgentDP}, M. for MapCoder, and O. for the original large model; Acc. for classification accuracy, Pre. for precision, Rec. for recall rate, and F1 for F1 score; the horizontal lines represent the user requirement (U.) for model deployment).}
    \label{fig:image_recognition_overall}
\vspace{-8pt}
\end{figure*}

\subsection{Metrics}
\label{sec:metrics}
We adopt three types of metrics to evaluate the real-time performance of the deployed models during inference:
1) \textbf{Resource Consumptions}. We measure the maximum \textit{RAM usage}, the maximum \textit{GPU memory usage}, and the \textit{model size}. Model size refers to the amount of disk space required to store the parameters of the embedded model.
2) \textbf{Task Accuracy}. For IR and SSA, we record the \textit{classification accuracy}, \textit{precision}, \textit{recall rate}, and \textit{F1 score} to assess how effectively the on-device model recognizes objects from input images or identifies emotions from audio data. For VLU, we measure the \textit{micro-averaged accuracy} as defined by the MMMU benchmark itself. 
3) \textbf{Real-time Performance}. For IR and SSA, we monitor the \textit{throughput}, \ie, the number of processed data samples per second. For VLU, we compute the \textit{tokens per second} (TPS) to evaluate the token generation speed of the embedded VLM. (4) \textbf{Wall-clock time.} To measure the overall efficiency of \textit{AgentDP}, we record the \textit{wall-clock time} from the moment a user specifies her requirements to the generation of the final embedded model deployment results.

\subsection{Baselines}
\label{sec:baseline}
\textbf{To the best of our knowledge, no existing system in literature can automatically execute the O\&A process for heterogeneous embedded devices} \cite{shen2026automating}. We select the best representative baselines, including: 1) \textbf{Original Models (O.)}, \ie, ResNet-152, the fine-tuned Whisper, and DeepSeek-VL-1.3b. 2) \textbf{MapCoder} \cite{islam2024mapcoder}, a SOTA multi-agent system tailored for complex programming tasks. We prompt MapCoder to synthesize programs for model optimization and adaptation, which are then manually executed for model deployment. 3) \textbf{Human baselines}. We invite 15 developers and evenly divide them into three groups based on their self-assessed development experience and expertise. \textbf{User Group 1 (U1)} are experts with extensive experience in deploying open-source models on various embedded devices. \textbf{User Group 2 (U2)} are semi-experts who only have extensive experience in training and optimizing AI models but with basic knowledge about embedded devices and model deployment. \textbf{User Group 3 (U3)} are non-experts who have limited knowledge of AI models and embedded devices with little practical experience. The evaluation process was reviewed and approved by our institution's ethics review board.

\subsection{Image Recognition Evaluation}
\subsubsection{\textbf{System Profiles \& User Requirements}}
Fig.\ref{fig:image_recognition_system_profile} illustrates the system profiles, where the original ResNet-152 model requires around 3.3 GB of RAM, with a model size of 230 MB. After removing ambiguities, the clarified user requirement (before standardization) is:

\begin{myquote}
  To deploy the ResNet-152 model on a Raspberry Pi 5, the RAM usage should stay below 3.12 GB. The model parameters should take up less than 200 MB of disk space. The system should process at least 5 images per second while maintaining a top-1 accuracy $\ge$ 65\%, precision $\ge$ 0.4, recall rate $\ge$ 0.5, and F1-score $\ge$ 0.45.
\end{myquote}

\subsubsection{\textbf{Overall Performance}}
Due to large hyperparameter search spaces, \textit{AgentDP} may generate optimized models with distinct performance.
To accommodate such randomness, we repeat the evaluation 20 times and obtain a set of different optimized models.
We then measure the metrics defined in \S~\ref{sec:metrics} and report the average values with their variances. From Fig.~\ref{fig:image_recognition_overall}, \textbf{we observe that:} 1) The model deployed by \textit{AgentDP} achieves a higher accuracy (over 13.8\% on average) while consuming fewer resources (reducing RAM usage by 14.5\% on average). 2) \textit{AgentDP} exhibits more consistent performance with smaller variances, whereas human baselines exhibit substantial fluctuations, especially in wall-clock time. 3) \textit{AgentDP} significantly reduces the time required for executing the O\&A process. Specifically, \textit{AgentDP} takes only 3.3 hours in total, with 2.8 hours spent on hyperparameter tuning, 0.3 hours on model adaptation, and 0.2 hours on information retrieval (\eg, searching). In contrast, even experts like U1 took substantial time (around 6 hours) on hyperparameter tuning to find a trade-off between model accuracy and resource consumption. 4) The model deployed by \textit{AgentDP} can sometimes achieve higher accuracy with fewer resources and higher throughput. \textit{These results highlight the exceptional efficiency of \textit{AgentDP} in finding optimal hyperparameters and generating high-performance optimized models for deployment. Such advances stem from our efficient multi-agent collaboration framework that fully automates the entire O\&A process using enhanced domain knowledge retrieved by various third-party tools.}

\begin{figure*}[t]
    \centering
    \subfigure[System profiles (simplified)]{
        \includegraphics[width=0.225\textwidth]{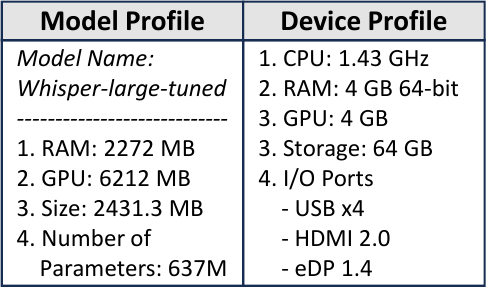}
        \label{fig:sentiment_analysis_system_profile}
    }
    \subfigure[Resource consumptions]{
        \includegraphics[width=0.225\textwidth]{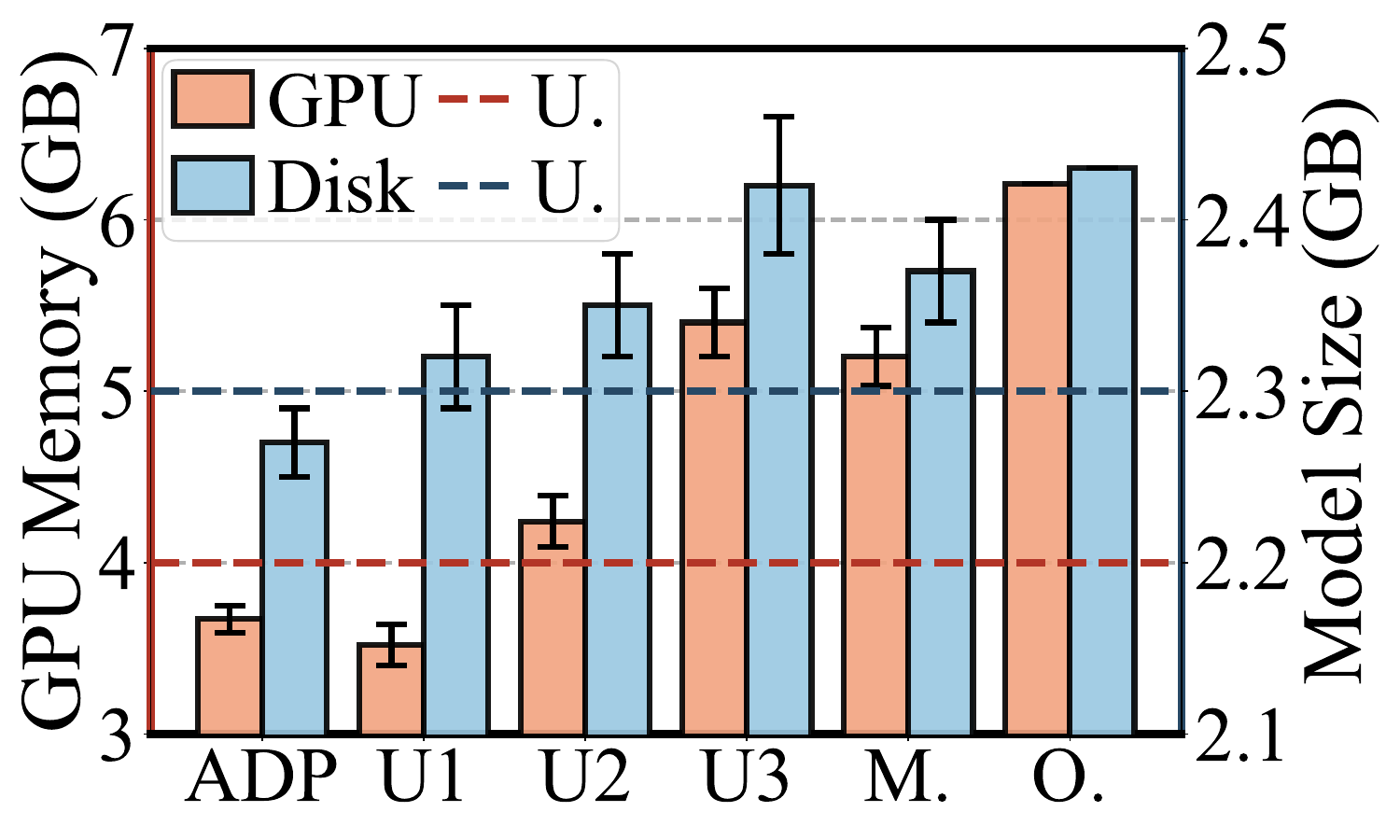}
        \label{fig:sentiment_analysis_resource_consumption}
    }
    \centering
    \subfigure[Task accuracy]{
        \includegraphics[width=0.225\textwidth]{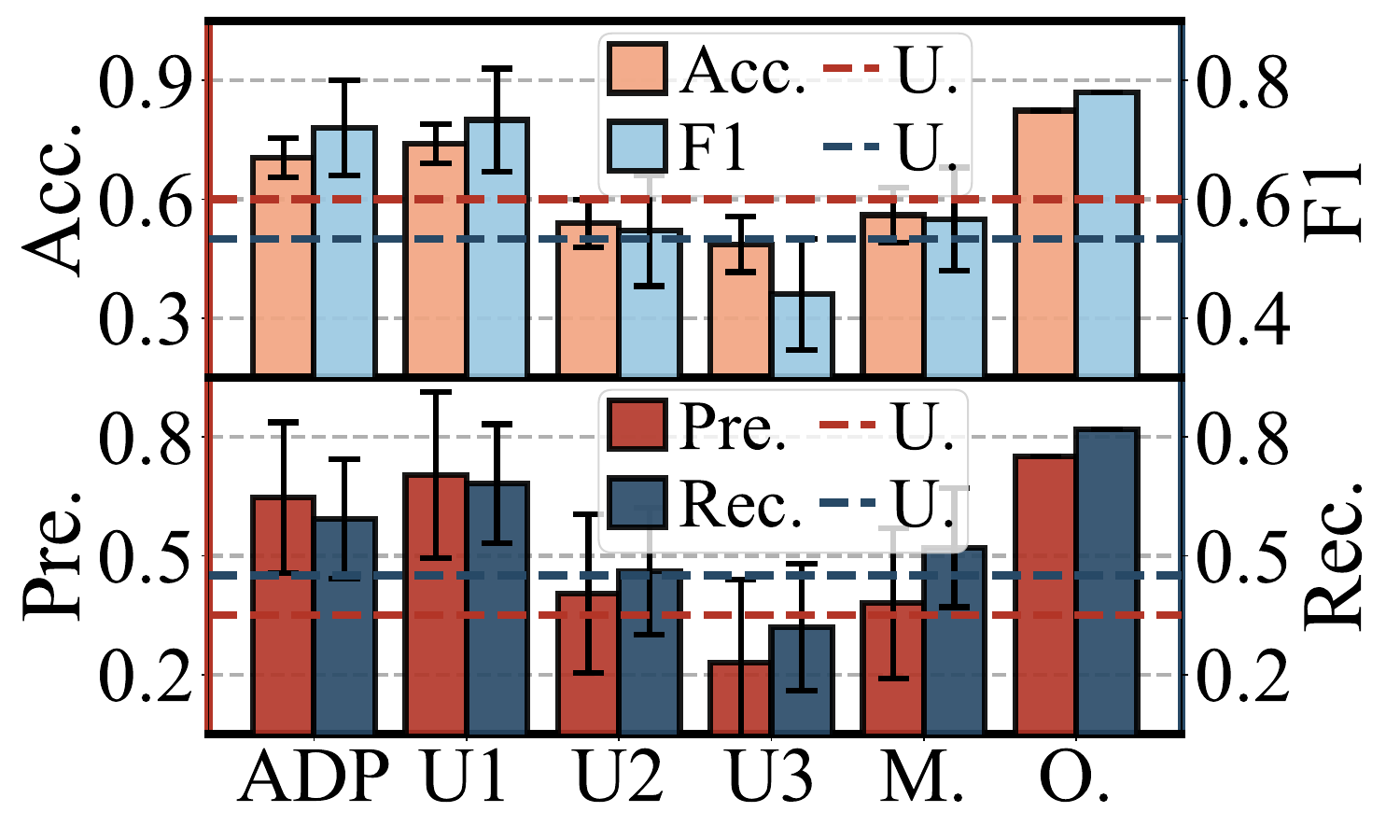}
        \label{fig:sentiment_analysis_task_accuracy}
    }
    \subfigure[Real-time performance]{
        \includegraphics[width=0.225\textwidth]{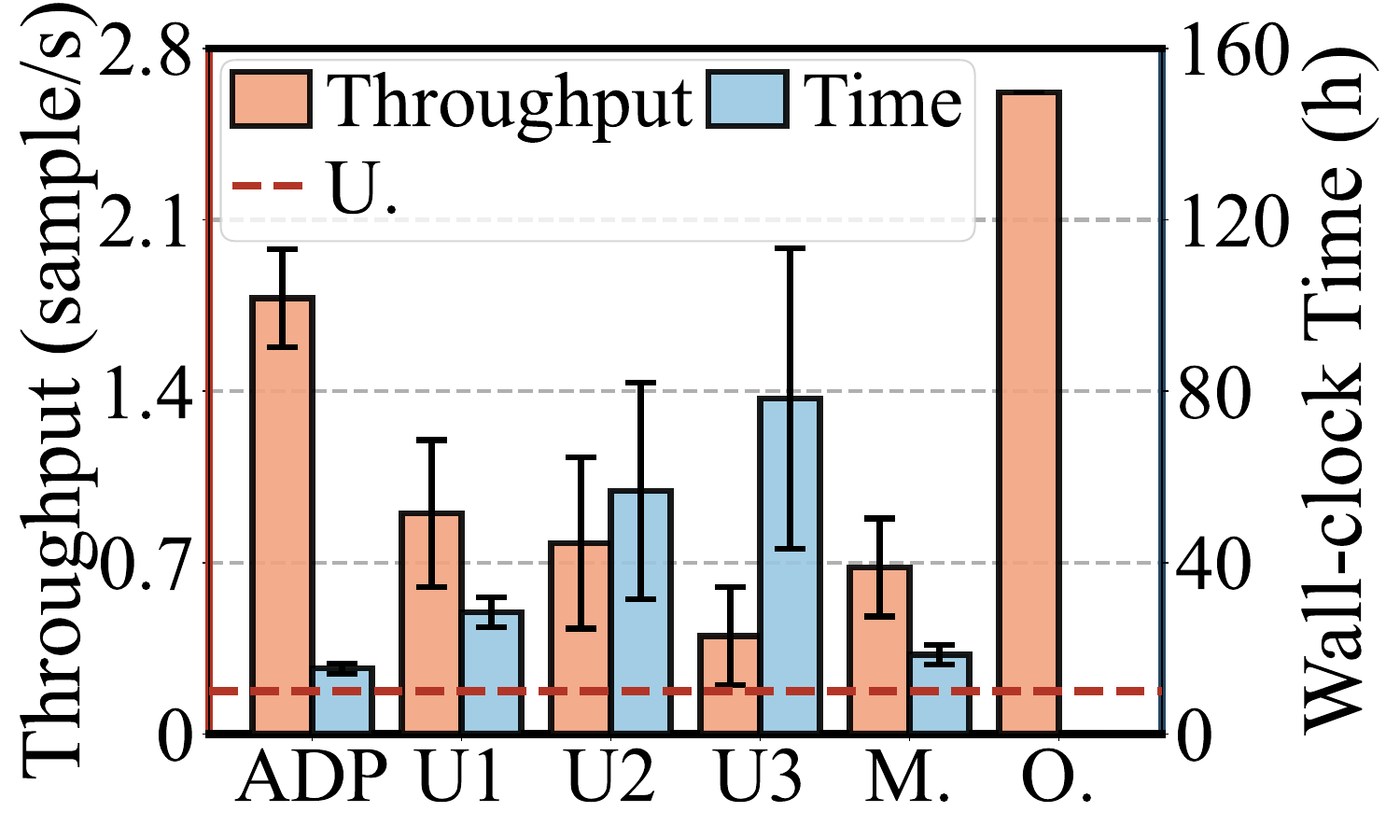}
        \label{fig:sentiment_analysis_realtime_performance}
    }
    \caption{Evaluation results for speech sentiment analysis (SSA).}
    \label{fig:sentiment_analysis_overall}
    \vspace{-8pt}
\end{figure*}

\begin{figure*}[t]
    \centering
    \subfigure[System profiles (simplified)]{
        \includegraphics[width=0.225\textwidth]{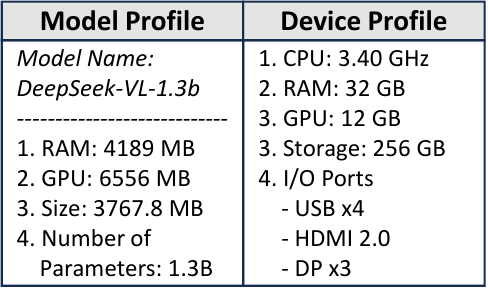}
        \label{fig:vision_language_system_profile}
    }
    \subfigure[Resource consumptions]{
        \includegraphics[width=0.225\textwidth]{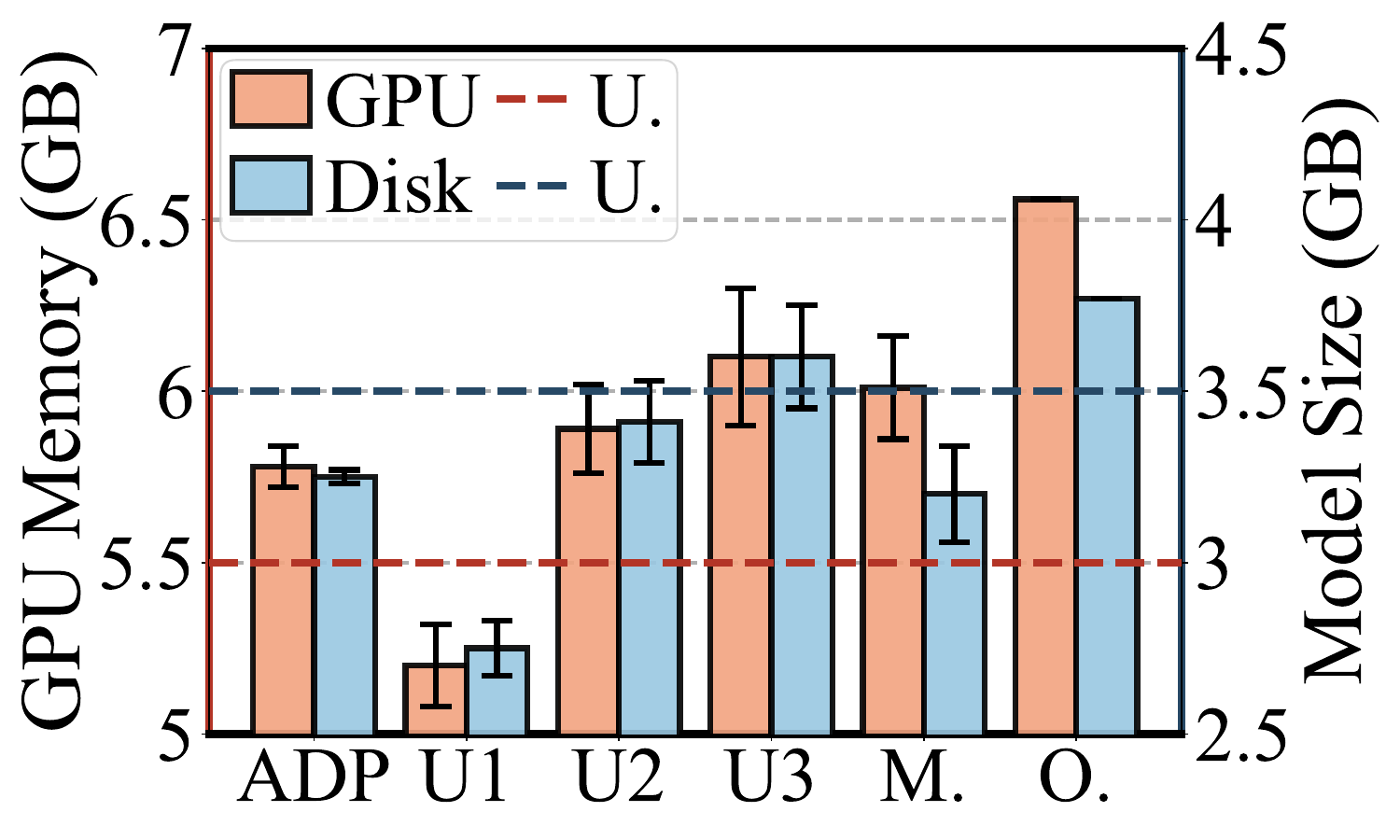}
        \label{fig:vision_language_resource_consumption}
    }
    \centering
    \subfigure[Task accuracy]{
        \includegraphics[width=0.225\textwidth]{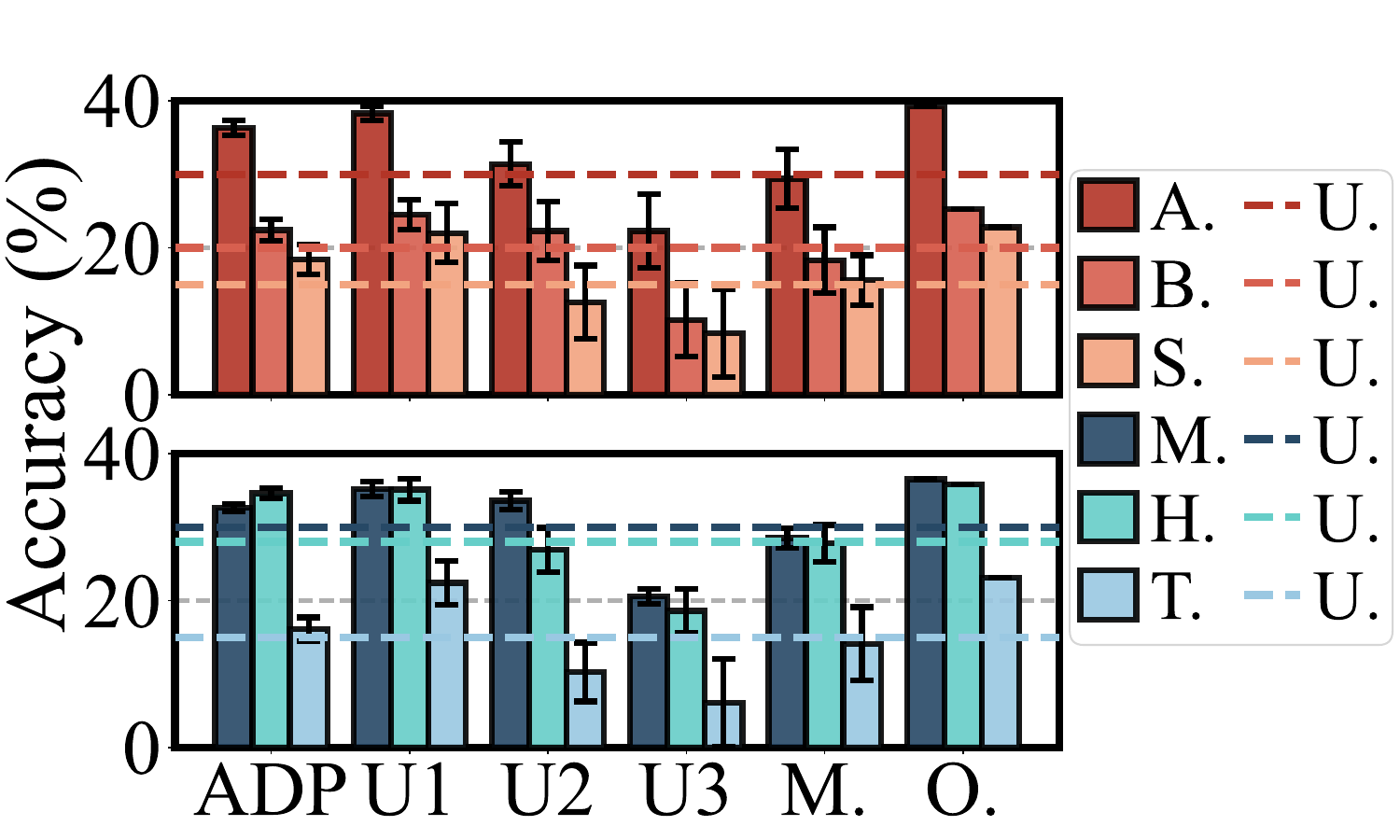}
        \label{fig:vision_language_task_accuracy}
    }
    \subfigure[Real-time performance]{
        \includegraphics[width=0.225\textwidth]{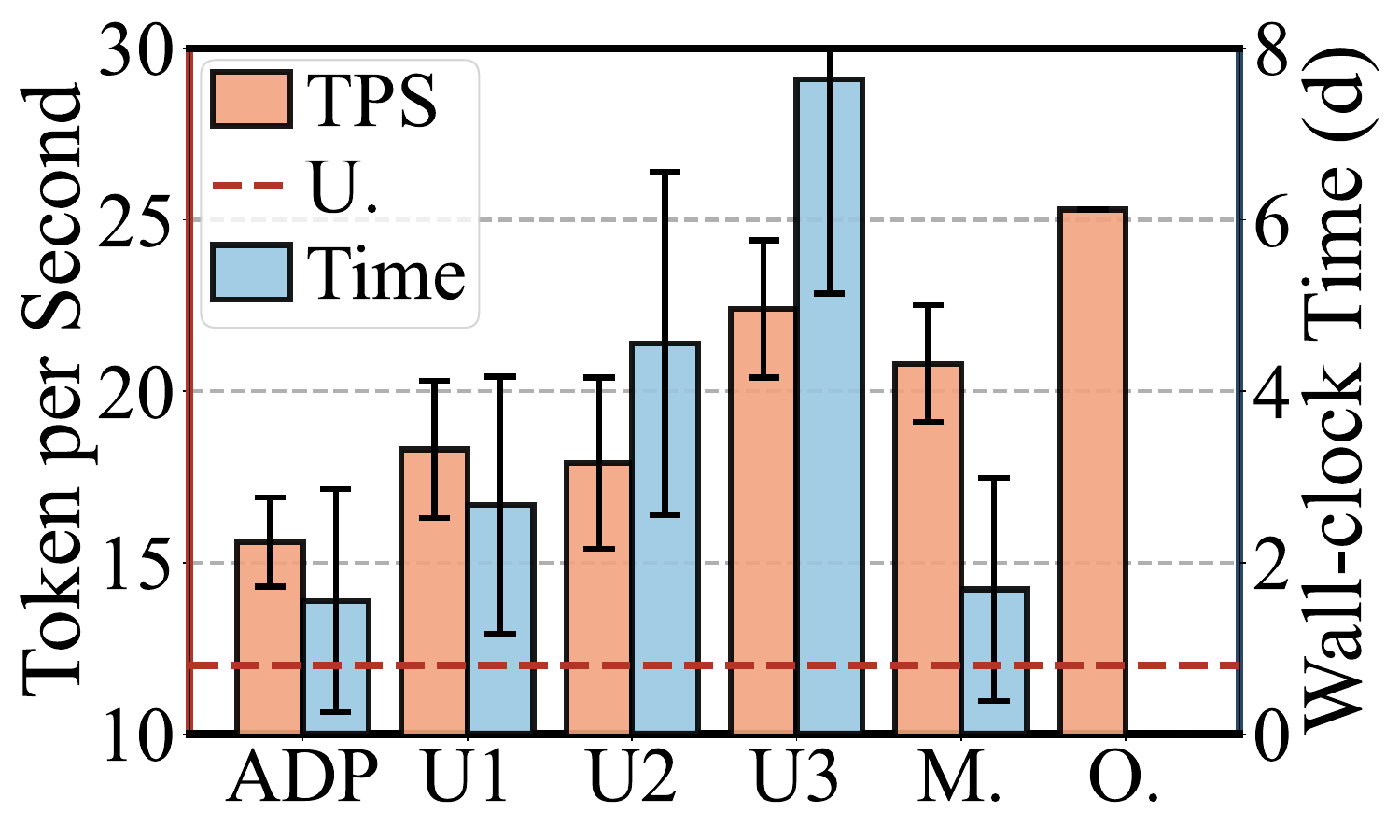}
        \label{fig:vision_language_realtime_performance}
    }
    \caption{Evaluation results for vision-language understanding (In (c), A. for art \& design, B. for business, S. for science, M. for medicine \& healthcare, H. for humanity, and T. for technology \& engineering).}
    \label{fig:vision_language_understanding_overall}
    \vspace{-8pt}
\end{figure*}

\subsection{Speech Sentiment Analysis Evaluation}
\subsubsection{\textbf{System Profiles \& User Requirements}}
\label{sec:user_requirement}
As shown in Fig.~\ref{fig:sentiment_analysis_system_profile}, the Whisper model (fine-tuned on several audio data) needs around 6.2 GB of GPU memory, significantly exceeding the maximum GPU memory of a Jetson Nano (4 GB). The user requirement before standardization is:
\begin{myquote}
  To deploy Whisper-large-v3 on Jetson Nano, the model should consume less than 4 GB of GPU memory. The model parameters should take up less than 2.3 GB of disk space. The system should process at least one audio sample per second with a top-1 accuracy $\ge$ 60\%, precision $\ge$ 0.35, recall rate $\ge$ 0.45, and F1-score $\ge$ 0.4.
\end{myquote}

\subsubsection{\textbf{Overall Performance}}
SSA using the fine-tuned Whisper model is more complex than image recognition due to the model's larger size and more complex operators. Fig.~\ref{fig:sentiment_analysis_overall} shows the overall performance of SSA. \textbf{We have four key findings}: 1) Only \textit{AgentDP} satisfies both resource and performance requirements, while the baselines fall short in different aspects. For example, to achieve high task accuracy, U1 sacrifices more disk space to store model parameters. Besides, to save GPU memory, U1 adopts a smaller batch size during inference, leading to a much lower throughput (49\% less than \textit{AgentDP}). 2) Despite considerable effort, U2 and U3 fail to meet most user requirements due to the high complexity of the Whisper model and their lack of expertise in model optimization. On the contrary, we analyze the model optimization code generated by \textit{AgentDP} and find that it even adopts a pruning method tailored for Transformer-based models \cite{kwon2022fast}. 3) Although MapCoder completes the O\&A process in a comparable amount of time to \textit{AgentDP}, it delivers lower performance even than semi-experts while consuming more resources. This is because MapCoder is a general programming agent without specialized domain knowledge in embedded AI and thereby cannot effectively handle specific O\&A tasks. 4) \textit{AgentDP} takes significantly less time (17 hours) to finish the O\&A process, whereas U2 and U3 take about 60 hours and 80 hours, respectively. The underlying reasons are twofold. During model optimization, \textit{AgentDP} efficiently reuses previous hyperparameters to narrow down search spaces and accelerate model optimization while U2 and U3 rely primarily on brute-force search. During model adaptation, U2 and U3 repeatedly fail to deploy the optimized model due to mismatched operators and Jetson Nano's complex inference environment. As a result, they resort to executing the model on the CPU, leading to much lower inference speed and throughput. In contrast, \textit{AgentDP} partitions the model using our OAMP module and selectively switches between different engines via our CEIP module for efficient model inference. \textit{These results further showcase that \textit{AgentDP} can efficiently perform model optimization even for complex models while accommodating heterogeneous embedded AI environments with enhanced reliability.}

\subsection{VLU Evaluation}
\subsubsection{\textbf{System Profiles \& User Requirements}}
Fig.~\ref{fig:vision_language_system_profile} illustrates the system profiles for the Visual-language Understanding (VLU) application. The clarified user requirement before the standardization process is:

\begin{myquote}
  To deploy DeepSeek-VL-1.3b on the edge PC, the GPU memory usage should stay below 6 GB. The model size should be less than 3.5 GB. The system should generate responses at a speed of at least 12 tokens per second. For topics in art, business, science, medicine, humanities, and technology, the average accuracy should be 30\%, 20\%, 15\%, 30\%, 28\%, and 15\%, respectively.
\end{myquote}

\begin{figure}[t]
    \centering
    \subfigure[Ablation on URS]{
        \includegraphics[width=0.225\textwidth]{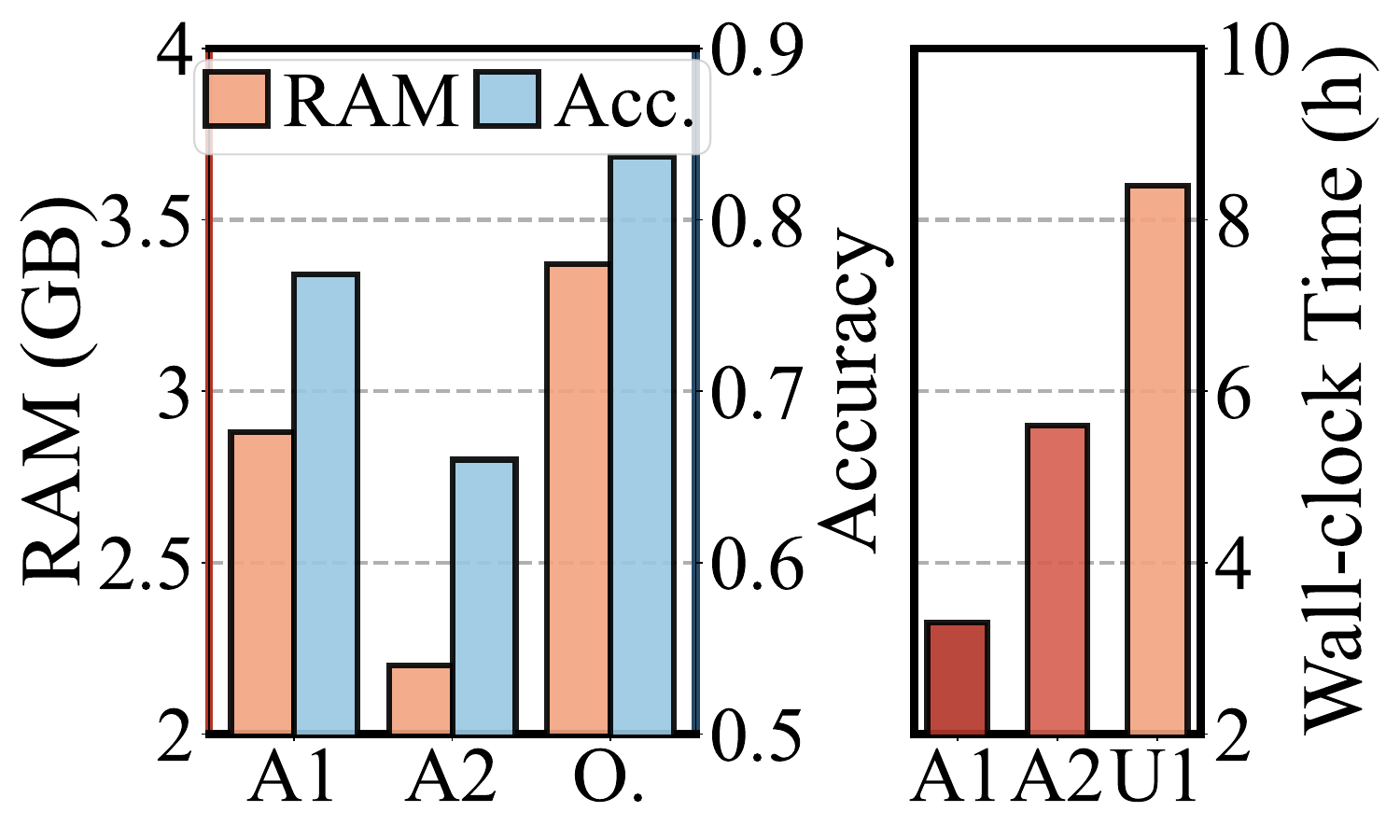}
        \label{fig:ablation_user_requirement_standardization}
    }
    \centering
    \subfigure[Ablation on configuration reuse]{
        \includegraphics[width=0.225\textwidth]{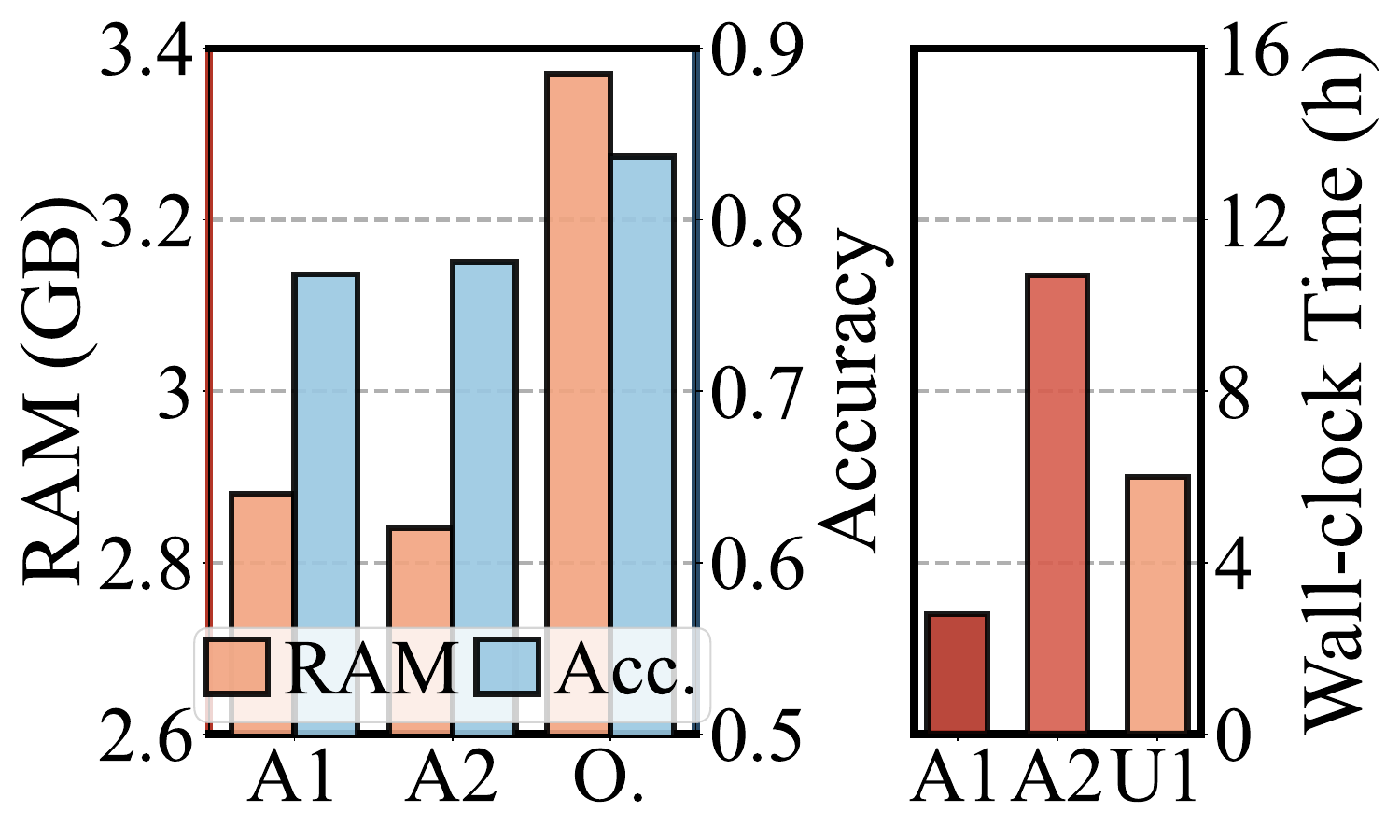}
        \label{fig:ablation_configuration_reuse}
    }
    \caption{Ablation study on image recognition (A1 for the original \textit{AgentDP}, A2 for \textit{AgentDP} without the URS module or CRA, and O. for the original large model).}
    \vspace{-8pt}
\end{figure}

\subsubsection{\textbf{Overall Performance}}
Fig.~\ref{fig:vision_language_understanding_overall} shows the overall performance of the VLM. \textbf{We find that}: 1) \textit{AgentDP}'s ability to reduce the time required for the O\&A process becomes more evident with larger models. Specifically, \textit{AgentDP} can achieve up to an 80\% reduction compared to human baselines. U2 and U3 even spend more than 4 days to finish the task due to their unfamiliarity with VLMs. 2) The overall accuracy of \textit{AgentDP} is slightly lower than that of U1. A deeper investigation reveals that U1 downloads several VLU datasets (\eg, MMBench \cite{liu2024mmbench}) and fine-tunes the optimized model to further enhance its performance. Thus, we can also add more tailored prompts to instruct \textit{AgentDP} to proactively adopt more advanced techniques (\eg, fine-tuning) to further improve the optimized model. \textit{These results exhibit the outstanding efficiency of \textit{AgentDP}, even when confronted with large models, stemming from our four fully automated and collaborative agents with enhanced domain expertise.}

\subsection{Ablation Study}
We conduct ablation studies to evaluate the impact of the technical modules in \textit{AgentDP}. For demonstration purposes, we use image recognition as an example, though similar phenomena are also observed in the other two applications.

\vspace{-2pt}
\subsubsection{\textbf{User Requirement Standardization}}
\label{sec:URS_ablation}
We directly input the raw user requirements into \textit{AgentDP} to execute the model O\&A process. As shown in Fig.~\ref{fig:ablation_user_requirement_standardization}, without our URS module, the deployed embedded model not only exhibits a significant classification accuracy drop (around 10\%) but also requires considerably more time to find a valid hyperparameter configuration. Further analysis of the source code generated by \textit{AgentDP} reveals that natural-language-described user requirements make it more difficult for HTA to interpret and verify each requirement. As a result, HTA creates fuzzy search spaces for hyperparameter tuning, resulting in prolonged optimization time. \textit{This underscores the importance of our URS module in helping \textit{AgentDP} gain a clear and comprehensive understanding of user requirements so that they can be effectively processed and verified.}

\subsubsection{\textbf{Configuration Reuse Agent.}}
\label{sec:ablation_reuse}
We first remove CRA by searching hyperparameters from scratch without reusing historical hyperparameters and then record the total time used for model optimization. As shown in Fig.~\ref{fig:ablation_configuration_reuse}, removing CRA dramatically increases the time required for model optimization (even exceeding expert users U1), while providing only a minor improvement in task accuracy and RAM usage. \textit{This implies that by reusing retrieved previous configurations for hyperparameter tuning, CRA enables \textit{AgentDP} to reduce the model optimization time by up to 74\% without searching from scratch as system utilization increases.}

\subsubsection{\textbf{OAMP \& CEIP}}
After removing our OAMP and CEIP module, we observe a significant decline in the token generation speed of the DeepSeek-VL model, dropping from 15.6 TPS to only 4.7 TPS. This reduction occurs because, without OAMP and CEIP, the portable model has to be executed on the CPU, as the model uses certain operators that are not fully supported by a single inference engine. \textit{This indicates that our OAMP and CEIP modules play a crucial role in improving model inference efficiency on embedded devices.}

\begin{table}[t]
\vspace{4pt}
\centering
\footnotesize
\caption{Participant information of user study (N=20)}
\label{tab:user_information}
\begin{tabular}{ll}
\toprule[1pt]
\textbf{Category}                             & \textbf{Background Information}                                                                 \\ \toprule[1pt]
\textbf{Gender}                               & Female (45\%), Male (40\%), Prefer not to say (15\%)\\ 
\textbf{Age}                                  & Under 18 (5\%), 18-30 (75\%), 30-39 (10\%), 40 and older (10\%)           \\ 
\textbf{Education}                      & Bachelor (5\%), Master (20\%), Doctoral (70\%), Others (5\%) \\ 
\textbf{English}     & Beginner (5\%), Intermediate (20\%), Advanced (65\%), Fluent (10\%)                              \\ 
\textbf{Expertise} & Expert (50\%), Semi-expert (50\%)               \\ \bottomrule[1pt]
\end{tabular}
\vspace{-4pt}
\end{table}

\subsection{User Study for Generalizability Evaluation}
To assess the utility, overall user satisfaction, and generalizability of \textit{AgentDP}, we conduct a user study involving 10 new expert users and 10 semi-expert users. \textit{\textbf{In order to test the generalizability of \textit{AgentDP}, we did not limit the tasks. Instead, participants are explicitly asked to freely express their requirements for deploying various open-source models on embedded devices to evaluate system generalizability.}} Expert users have rich experience in model optimization and have developed multiple embedded AI applications, while semi-expert users are proficient in model optimization but possess only basic knowledge of embedded devices. 
Table~\ref{tab:user_information} lists the detailed background information of the participants.

\subsubsection{\textbf{Metrics}}
We ask users to rate \textit{AgentDP} based on five metrics: 1) \textit{Overall Model Performance} \textbf{(OMP)} evaluates the overall performance of the deployed model on private data on the target device, considering resource consumption, task accuracy, and real-time performance; 2) \textit{Code Readability} \textbf{(CR)} measures the clarity and structure of the generated code for the O\&A process; 3) \textit{Efficiency of Model Optimization} \textbf{(EMO)} assesses how efficiently \textit{AgentDP} finds optimal hyperparameter sets for model optimization; 4) \textit{Efficiency of Model Deployment} \textbf{(EMD)} evaluates how efficiently \textit{AgentDP} adapts the optimized model to heterogeneous embedded devices for deployment; 5) \textit{User Satisfaction} \textbf{(US)} captures users' overall experience with \textit{AgentDP} during usage. All metrics are rated on a scale from 1 (not at all) to 5 (completely).

\subsubsection{\textbf{Overall Performance}}
\begin{figure}[t]
    \centering
    \subfigure[Overall performance]{
        \includegraphics[width=0.225\textwidth]{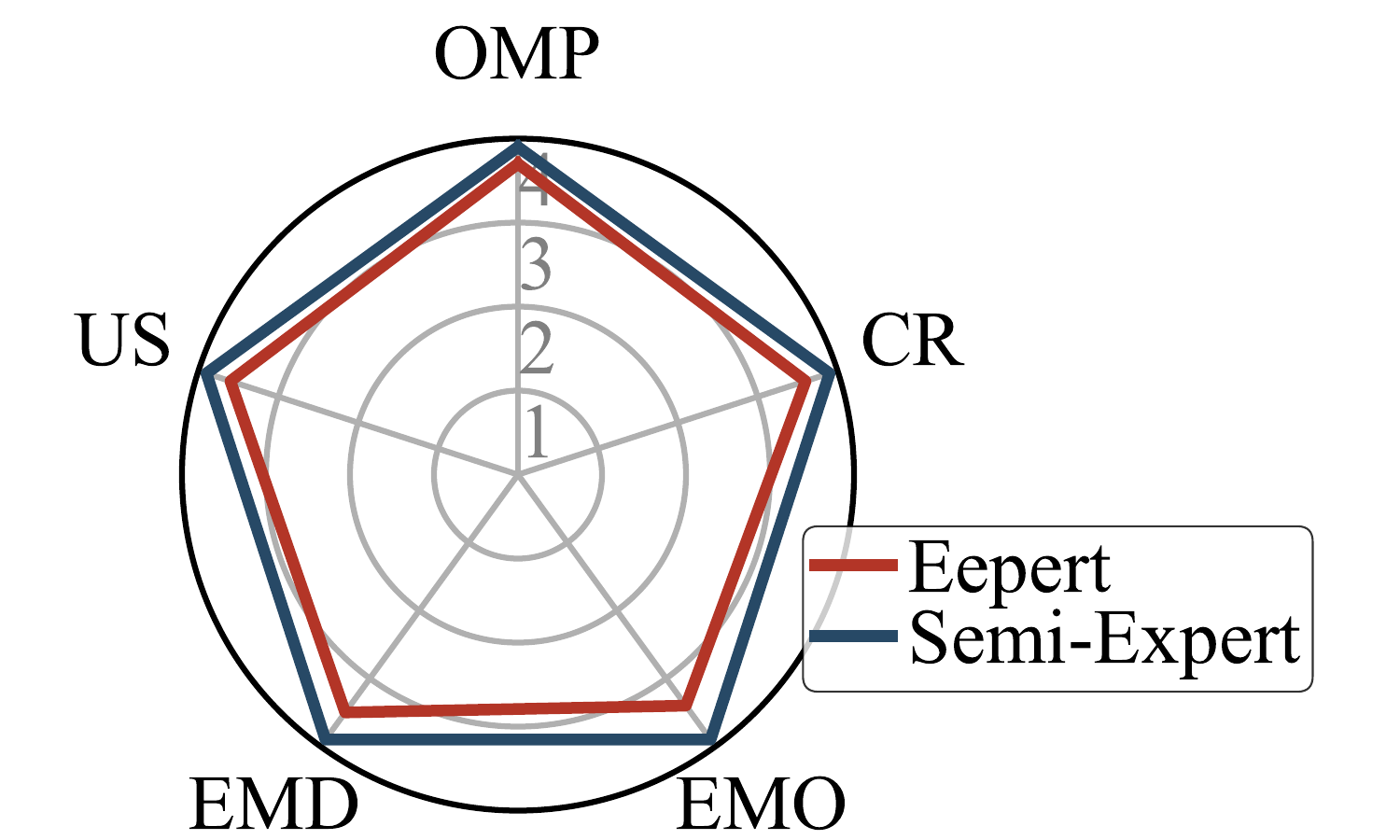}
        \label{fig:user_study_overall_performance}
    }
    \centering
    \subfigure[Long-term evaluation]{
        \includegraphics[width=0.225\textwidth]{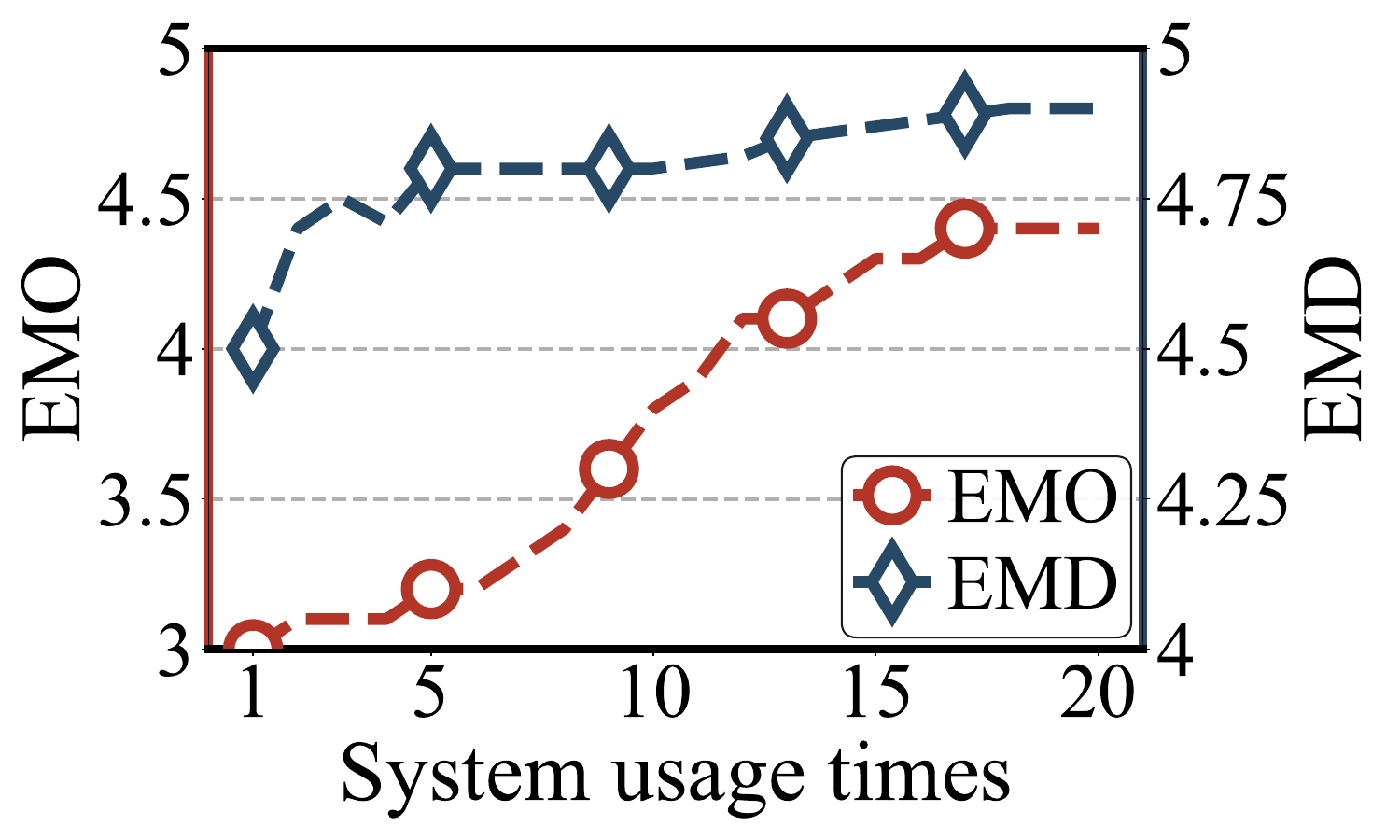}
        \label{fig:user_study_long_term}
    }
    \caption{User study.}
    \vspace{-8pt}
\end{figure}

Fig.~\ref{fig:user_study_overall_performance} shows the overall performance of the user study with averaged scores. Both experts and semi-experts provide high ratings (above 4.6) for \textit{AgentDP} in terms of OMP, CR, and US. However, semi-expert users tend to give higher scores than expert users for EMO and EMD. This discrepancy stems from semi-experts' lack of extensive experience with various model optimization strategies and execution environment configurations tailored for heterogeneous embedded devices. As a result, they tend to under-specify their requirements and may occasionally introduce conflicting conditions. Thus, they are more likely to rate higher scores for optimization efficiency since \textit{AgentDP} can effectively determine appropriate search spaces for hyperparameter tuning and automatically adapt portable models to the target device using device-specific inference engines. \textit{Such results exhibit \textit{AgentDP}'s high generalizability in adapting various open-source models to heterogeneous embedded devices with enhanced user experience.}

\subsubsection{\textbf{Long-term Evaluation}}
We also invite expert users to participate in a long-term evaluation to investigate the impact on EMO and EMD as system utilization increases. Fig.~\ref{fig:user_study_long_term} shows that with increasing user engagement, both EMO and EMD gradually increase, with EMO exhibiting faster growth. As system usage time increases, CRA stores more historical hyperparameter configurations in the graph database and can retrieve similar ones for reuse when new user requests arrive. Consequently, the search spaces for hyperparameter tuning gradually narrow, reducing the time required for finding optimal hyperparameter sets tailored for the resource-constrained target device, leading to a higher EMO score. Additionally, previous model profiles and device profiles can also be reused during deployment, resulting in increased EMD scores. \textit{These results demonstrate the outstanding usability and efficiency of \textit{AgentDP} in long-term use.}

\section{Discussion}

\noindent\textbf{Error Handling in Multi-Agent Systems.}
In LLM agent systems, a common challenge is the generation of incorrect intermediate results during multi-step reasoning. These errors can propagate and lead to incorrect final outputs, significantly impacting the reliability of the agent. To address this, AgentDP incorporates Reflexion \cite{shinn2023reflexion} that enables agents to reflect on failed attempts and iteratively improve. \textit{Moreover, to further enhance the robustness of \textit{AgentDP} in helping agents detect and recover from mistakes during multi-step reasoning, we plan to orthogonally integrate more advanced techniques (\eg, ReAct \cite{yao2023react}) as part of our future work.}

\noindent\textbf{Decoupled HTA and MAA.} HTA primarily focuses on reducing the resource consumption of large models via hyperparameter tuning, ensuring that embedded devices can accommodate them. The real-time performance (\eg, inference latency), however, is only recorded during model deployment on embedded devices. Thus, in multiple rounds of O\&A optimization, HTA can only adjust hyperparameters for further optimization based on MAA's feedback and present users with all applicable hyperparameter configurations that fully satisfy their requirements. To further enhance the efficiency in such an iterative process, we can orthogonally integrate real-time performance prediction models \cite{zhang2021nn} with \textit{AgentDP} to narrow down the search space during hyperparameter tuning.

\section{Related Work}

\noindent\textbf{Model Deployment on Embedded Devices.}
Most pre-trained models cannot be directly deployed on embedded devices due to limited resources and heterogeneous execution environments \cite{park2024fedhm}. Existing works \cite{liu2024dynaspa, jia2024tinytta} on model deployment to embedded devices focus on optimizing deep learning models for limited compute, memory, and energy budgets.
However, hyperparameter tuning often requires multiple iterations of trial-and-error, resulting in substantial human effort even with SOTA hyperparameter tuning frameworks \cite{akiba2019optuna}.
On the other hand, while numerous model deployment frameworks \cite{onnxruntime} exist, they are not tailored for heterogeneous embedded environments. Developers need to first generate portable models and deploy them on device-specific inference engines (\eg, ONNX Runtime \cite{onnxruntime}). Mastering such highly specialized knowledge poses huge challenges even for experienced experts. To deal with such heterogeneous environments, existing solutions either rely on fallback environments (\eg, CPU) for model execution or continuously develop new cross-platform frameworks (\eg, ExecuTorch \cite{torchedge2022executorch}), both of which are inefficient and cumbersome. In addition, although container techniques (\eg, Docker \cite{merkel2014docker}) can package dependencies to ease software portability, they often fail to support heterogeneous hardware architectures in embedded AI scenarios.
To address these issues, instead of requiring developers to manually configure the model optimization and adaptation processes, \textit{AgentDP} fully automates the whole O\&A process with enriched domain knowledge, significantly reducing human effort with enhanced reliability.

\noindent\textbf{LLM Agents.}
LLM agents \cite{shen2025autoiot} automate complex tasks by integrating LLMs with various third-party tools (\eg, web search engines).
These agents can be applied to various applications, such as program synthesis \cite{shen2025gpiot, luo2024repoagent} and smart assistants \cite{xu2024can}. Furthermore, some studies \cite{huang2023agentcoder, islam2024mapcoder} construct multiple agents to collaboratively accomplish complex tasks, allowing specialized agents to work together and adapt to dynamic environments efficiently. For example, MapCoder \cite{islam2024mapcoder} solves competitive programming tasks via four agents that handle different development stages.
Unlike Code LLMs like MapCoder, which are designed to synthesize general programs, \textit{AgentDP} builds an automation chain to optimize and adapt open-source models for heterogeneous embedded devices, addressing unique technical challenges (\eg, model-engine mismatch) and significantly reducing human effort with enhanced model performance.


\section{Conclusion}

This paper presents \textit{AgentDP}, a multi-agent automatic model optimization and adaptation (O\&A) system for heterogeneous embedded AI applications. Armed with four specialized agents to address three distinct technical challenges, \textit{AgentDP} significantly enhances the O\&A efficiency with enhanced model performance. User studies demonstrate \textit{AgentDP}'s high generalizability in adapting various open-source models to heterogeneous embedded devices with distinct user requirements. We believe \textit{AgentDP} can unlock the potential to drive the practical deployment of open-source AI models across diverse and distributed embedded devices in real-world everyday applications.



\clearpage
\balance
\bibliographystyle{ACM-Reference-Format}

\begin{thebibliography}{72}


\ifx \showCODEN    \undefined \def \showCODEN     #1{\unskip}     \fi
\ifx \showISBNx    \undefined \def \showISBNx     #1{\unskip}     \fi
\ifx \showISBNxiii \undefined \def \showISBNxiii  #1{\unskip}     \fi
\ifx \showISSN     \undefined \def \showISSN      #1{\unskip}     \fi
\ifx \showLCCN     \undefined \def \showLCCN      #1{\unskip}     \fi
\ifx \shownote     \undefined \def \shownote      #1{#1}          \fi
\ifx \showarticletitle \undefined \def \showarticletitle #1{#1}   \fi
\ifx \showURL      \undefined \def \showURL       {\relax}        \fi
\providecommand\bibfield[2]{#2}
\providecommand\bibinfo[2]{#2}
\providecommand\natexlab[1]{#1}
\providecommand\showeprint[2][]{arXiv:#2}

\bibitem[onn(2021a)]%
        {onnxruntime}
 \bibinfo{year}{2021}\natexlab{a}.
\newblock \bibinfo{title}{ONNX Runtime}.
\newblock \bibinfo{howpublished}{\url{https://onnxruntime.ai/}}.
\newblock


\bibitem[onn(2021b)]%
        {onnx}
 \bibinfo{year}{2021}\natexlab{b}.
\newblock \bibinfo{title}{Open Neural Network Exchange}.
\newblock \bibinfo{howpublished}{\url{https://onnx.ai/}}.
\newblock


\bibitem[Bas(2022)]%
        {Bashir2022chroma}
 \bibinfo{year}{2022}\natexlab{}.
\newblock \bibinfo{title}{Chroma}.
\newblock
\urldef\tempurl%
\url{https://github.com/chroma-core/chroma}
\showURL{%
\tempurl}


\bibitem[Cam(2022)]%
        {Camara2022firecrawl}
 \bibinfo{year}{2022}\natexlab{}.
\newblock \bibinfo{title}{Firecrawl}.
\newblock
\urldef\tempurl%
\url{https://github.com/mendableai/firecrawl}
\showURL{%
\tempurl}


\bibitem[jet(2022)]%
        {jetson_nano_configuration}
 \bibinfo{year}{2022}\natexlab{}.
\newblock \bibinfo{title}{Get Started With Jetson Nano Developer Kit}.
\newblock \bibinfo{howpublished}{\url{https://developer.nvidia.com/embedded/learn/get-started-jetson-nano-devkit}}.
\newblock


\bibitem[Fin(2022)]%
        {Finne2022neo4j}
 \bibinfo{year}{2022}\natexlab{}.
\newblock \bibinfo{title}{Neo4j}.
\newblock
\urldef\tempurl%
\url{https://github.com/neo4j/neo4j}
\showURL{%
\tempurl}


\bibitem[ope(2022)]%
        {openai2022textembeddingada}
 \bibinfo{year}{2022}\natexlab{}.
\newblock \bibinfo{title}{OpenAI text-embedding-ada-002}.
\newblock \bibinfo{howpublished}{\url{https://platform.openai.com/docs/guides/embeddings/embedding-models}}.
\newblock


\bibitem[Ash(2022)]%
        {Asher2022tavily}
 \bibinfo{year}{2022}\natexlab{}.
\newblock \bibinfo{title}{Tavily}.
\newblock
\urldef\tempurl%
\url{https://github.com/tavily-ai/tavily-python}
\showURL{%
\tempurl}


\bibitem[ope(2023)]%
        {openvino}
 \bibinfo{year}{2023}\natexlab{}.
\newblock \bibinfo{title}{OpenVINO}.
\newblock \bibinfo{howpublished}{\url{https://github.com/openvinotoolkit/openvino}}.
\newblock


\bibitem[ten(2024)]%
        {tensorrt_mod_operator}
 \bibinfo{year}{2024}\natexlab{}.
\newblock
\urldef\tempurl%
\url{https://github.com/onnx/onnx-tensorrt/blob/84b5be1d6fc03564f2c0dba85a2ee75bad242c2e/operators.md}
\showURL{%
\tempurl}


\bibitem[onn(2024)]%
        {onnxruntime_mod_operator}
 \bibinfo{year}{2024}\natexlab{}.
\newblock
\urldef\tempurl%
\url{https://onnx.ai/onnx/operators/onnx__Mod.html}
\showURL{%
\tempurl}


\bibitem[doc(2025)]%
        {docker_container_architecture}
 \bibinfo{year}{2025}\natexlab{}.
\newblock
\urldef\tempurl%
\url{https://docs.docker.com/build/building/multi-platform/}
\showURL{%
\tempurl}


\bibitem[Abraham et~al\mbox{.}(2023)]%
        {abraham2023homopt}
\bibfield{author}{\bibinfo{person}{Sophia~J Abraham}, \bibinfo{person}{Kehelwala~DG Maduranga}, \bibinfo{person}{Jeffery Kinnison}, \bibinfo{person}{Zachariah Carmichael}, \bibinfo{person}{Jonathan~D Hauenstein}, {and} \bibinfo{person}{Walter~J Scheirer}.} \bibinfo{year}{2023}\natexlab{}.
\newblock \showarticletitle{Homopt: A homotopy-based hyperparameter optimization method}.
\newblock \bibinfo{journal}{\emph{arXiv preprint arXiv:2308.03317}} (\bibinfo{year}{2023}).
\newblock


\bibitem[Achiam et~al\mbox{.}(2023)]%
        {achiam2023gpt}
\bibfield{author}{\bibinfo{person}{Josh Achiam}, \bibinfo{person}{Steven Adler}, \bibinfo{person}{Sandhini Agarwal}, \bibinfo{person}{Lama Ahmad}, \bibinfo{person}{Ilge Akkaya}, \bibinfo{person}{Florencia~Leoni Aleman}, \bibinfo{person}{Diogo Almeida}, \bibinfo{person}{Janko Altenschmidt}, \bibinfo{person}{Sam Altman}, \bibinfo{person}{Shyamal Anadkat}, {et~al\mbox{.}}} \bibinfo{year}{2023}\natexlab{}.
\newblock \showarticletitle{Gpt-4 technical report}.
\newblock \bibinfo{journal}{\emph{arXiv preprint arXiv:2303.08774}} (\bibinfo{year}{2023}).
\newblock


\bibitem[Ahn et~al\mbox{.}(2023)]%
        {ahn2023performance}
\bibfield{author}{\bibinfo{person}{Hyunho Ahn}, \bibinfo{person}{Tian Chen}, \bibinfo{person}{Nawras Alnaasan}, \bibinfo{person}{Aamir Shafi}, \bibinfo{person}{Mustafa Abduljabbar}, \bibinfo{person}{Hari Subramoni}, {et~al\mbox{.}}} \bibinfo{year}{2023}\natexlab{}.
\newblock \showarticletitle{Performance characterization of using quantization for dnn inference on edge devices: Extended version}.
\newblock \bibinfo{journal}{\emph{arXiv preprint arXiv:2303.05016}} (\bibinfo{year}{2023}).
\newblock


\bibitem[Akiba et~al\mbox{.}(2019)]%
        {akiba2019optuna}
\bibfield{author}{\bibinfo{person}{Takuya Akiba}, \bibinfo{person}{Shotaro Sano}, \bibinfo{person}{Toshihiko Yanase}, \bibinfo{person}{Takeru Ohta}, {and} \bibinfo{person}{Masanori Koyama}.} \bibinfo{year}{2019}\natexlab{}.
\newblock \showarticletitle{Optuna: A next-generation hyperparameter optimization framework}. In \bibinfo{booktitle}{\emph{{ACM} {KDD}}}.
\newblock


\bibitem[Brown et~al\mbox{.}(2020)]%
        {brown2020language}
\bibfield{author}{\bibinfo{person}{Tom Brown}, \bibinfo{person}{Benjamin Mann}, \bibinfo{person}{Nick Ryder}, \bibinfo{person}{Melanie Subbiah}, \bibinfo{person}{Jared~D Kaplan}, \bibinfo{person}{Prafulla Dhariwal}, \bibinfo{person}{Arvind Neelakantan}, \bibinfo{person}{Pranav Shyam}, \bibinfo{person}{Girish Sastry}, \bibinfo{person}{Amanda Askell}, {et~al\mbox{.}}} \bibinfo{year}{2020}\natexlab{}.
\newblock \showarticletitle{Language models are few-shot learners}.
\newblock \bibinfo{journal}{\emph{{NeurIPS}}} (\bibinfo{year}{2020}).
\newblock


\bibitem[Chase et~al\mbox{.}(2022)]%
        {Chase2022LangChain}
\bibfield{author}{\bibinfo{person}{Harrison Chase} {et~al\mbox{.}}} \bibinfo{year}{2022}\natexlab{}.
\newblock \bibinfo{title}{LangChain: Build context-aware reasoning applications}.
\newblock
\urldef\tempurl%
\url{https://github.com/langchain-ai/langchain}
\showURL{%
\tempurl}


\bibitem[Chen et~al\mbox{.}(2018)]%
        {chen2018tvm}
\bibfield{author}{\bibinfo{person}{Tianqi Chen}, \bibinfo{person}{Thierry Moreau}, \bibinfo{person}{Ziheng Jiang}, \bibinfo{person}{Lianmin Zheng}, \bibinfo{person}{Eddie Yan}, \bibinfo{person}{Haichen Shen}, \bibinfo{person}{Meghan Cowan}, \bibinfo{person}{Leyuan Wang}, \bibinfo{person}{Yuwei Hu}, \bibinfo{person}{Luis Ceze}, {et~al\mbox{.}}} \bibinfo{year}{2018}\natexlab{}.
\newblock \showarticletitle{$\{$TVM$\}$: An automated $\{$End-to-End$\}$ optimizing compiler for deep learning}. In \bibinfo{booktitle}{\emph{{USENIX} {OSDI}}}.
\newblock


\bibitem[Deng et~al\mbox{.}(2009)]%
        {deng2009imagenet}
\bibfield{author}{\bibinfo{person}{Jia Deng}, \bibinfo{person}{Wei Dong}, \bibinfo{person}{Richard Socher}, \bibinfo{person}{Li-Jia Li}, \bibinfo{person}{Kai Li}, {and} \bibinfo{person}{Li Fei-Fei}.} \bibinfo{year}{2009}\natexlab{}.
\newblock \showarticletitle{Imagenet: A large-scale hierarchical image database}. In \bibinfo{booktitle}{\emph{2009 IEEE conference on computer vision and pattern recognition}}. Ieee, \bibinfo{pages}{248--255}.
\newblock


\bibitem[development team(2022)]%
        {torchedge2022executorch}
\bibfield{author}{\bibinfo{person}{PyTorch development team}.} \bibinfo{year}{2022}\natexlab{}.
\newblock \bibinfo{title}{ExecuTorch: A powerful on-device AI Framework}.
\newblock \bibinfo{howpublished}{\url{https://github.com/pytorch/executorch}}.
\newblock


\bibitem[Falkner et~al\mbox{.}(2018)]%
        {falkner2018bohb}
\bibfield{author}{\bibinfo{person}{Stefan Falkner}, \bibinfo{person}{Aaron Klein}, {and} \bibinfo{person}{Frank Hutter}.} \bibinfo{year}{2018}\natexlab{}.
\newblock \showarticletitle{BOHB: Robust and efficient hyperparameter optimization at scale}. In \bibinfo{booktitle}{\emph{{ICML}}}.
\newblock


\bibitem[He et~al\mbox{.}(2017)]%
        {he2017mask}
\bibfield{author}{\bibinfo{person}{Kaiming He}, \bibinfo{person}{Georgia Gkioxari}, \bibinfo{person}{Piotr Doll{\'a}r}, {and} \bibinfo{person}{Ross Girshick}.} \bibinfo{year}{2017}\natexlab{}.
\newblock \showarticletitle{Mask r-cnn}. In \bibinfo{booktitle}{\emph{{IEEE} {ICCV}}}. \bibinfo{pages}{2961--2969}.
\newblock


\bibitem[He et~al\mbox{.}(2016)]%
        {he2016deep}
\bibfield{author}{\bibinfo{person}{Kaiming He}, \bibinfo{person}{Xiangyu Zhang}, \bibinfo{person}{Shaoqing Ren}, {and} \bibinfo{person}{Jian Sun}.} \bibinfo{year}{2016}\natexlab{}.
\newblock \showarticletitle{Deep residual learning for image recognition}. In \bibinfo{booktitle}{\emph{Proceedings of the IEEE conference on computer vision and pattern recognition}}. \bibinfo{pages}{770--778}.
\newblock


\bibitem[He et~al\mbox{.}(2023)]%
        {he2023vi}
\bibfield{author}{\bibinfo{person}{Yuze He}, \bibinfo{person}{Chen Bian}, \bibinfo{person}{Jingfei Xia}, \bibinfo{person}{Shuyao Shi}, \bibinfo{person}{Zhenyu Yan}, \bibinfo{person}{Qun Song}, {and} \bibinfo{person}{Guoliang Xing}.} \bibinfo{year}{2023}\natexlab{}.
\newblock \showarticletitle{Vi-map: Infrastructure-assisted real-time hd mapping for autonomous driving}. In \bibinfo{booktitle}{\emph{{ACM} {MobiCom}}}. \bibinfo{pages}{1--15}.
\newblock


\bibitem[Huang et~al\mbox{.}(2023)]%
        {huang2023agentcoder}
\bibfield{author}{\bibinfo{person}{Dong Huang}, \bibinfo{person}{Jie~M Zhang}, \bibinfo{person}{Michael Luck}, \bibinfo{person}{Qingwen Bu}, \bibinfo{person}{Yuhao Qing}, {and} \bibinfo{person}{Heming Cui}.} \bibinfo{year}{2023}\natexlab{}.
\newblock \showarticletitle{Agentcoder: Multi-agent-based code generation with iterative testing and optimisation}.
\newblock \bibinfo{journal}{\emph{arXiv preprint arXiv:2312.13010}} (\bibinfo{year}{2023}).
\newblock


\bibitem[Huang et~al\mbox{.}(2026)]%
        {huang2026jailbreaking}
\bibfield{author}{\bibinfo{person}{Xinyu Huang}, \bibinfo{person}{Qiang Yang}, \bibinfo{person}{Leming Shen}, \bibinfo{person}{Zijing Ma}, {and} \bibinfo{person}{Yuanqing Zheng}.} \bibinfo{year}{2026}\natexlab{}.
\newblock \showarticletitle{Jailbreaking Embodied LLMs via Action-Level Manipulation}. In \bibinfo{booktitle}{\emph{Proceedings of the 2026 ACM/IEEE International Conference on Embedded Artificial Intelligence and Sensing Systems}}. \bibinfo{pages}{1057--1071}.
\newblock


\bibitem[Hurst et~al\mbox{.}(2024)]%
        {hurst2024gpt}
\bibfield{author}{\bibinfo{person}{Aaron Hurst}, \bibinfo{person}{Adam Lerer}, \bibinfo{person}{Adam~P Goucher}, \bibinfo{person}{Adam Perelman}, \bibinfo{person}{Aditya Ramesh}, \bibinfo{person}{Aidan Clark}, \bibinfo{person}{AJ Ostrow}, \bibinfo{person}{Akila Welihinda}, \bibinfo{person}{Alan Hayes}, \bibinfo{person}{Alec Radford}, {et~al\mbox{.}}} \bibinfo{year}{2024}\natexlab{}.
\newblock \showarticletitle{Gpt-4o system card}.
\newblock \bibinfo{journal}{\emph{arXiv preprint arXiv:2410.21276}} (\bibinfo{year}{2024}).
\newblock


\bibitem[Ishibashi and Nishimura(2024)]%
        {ishibashi2024self}
\bibfield{author}{\bibinfo{person}{Yoichi Ishibashi} {and} \bibinfo{person}{Yoshimasa Nishimura}.} \bibinfo{year}{2024}\natexlab{}.
\newblock \showarticletitle{Self-organized agents: A llm multi-agent framework toward ultra large-scale code generation and optimization}.
\newblock \bibinfo{journal}{\emph{arXiv preprint arXiv:2404.02183}} (\bibinfo{year}{2024}).
\newblock


\bibitem[Islam et~al\mbox{.}(2024)]%
        {islam2024mapcoder}
\bibfield{author}{\bibinfo{person}{Md~Ashraful Islam}, \bibinfo{person}{Mohammed~Eunus Ali}, {and} \bibinfo{person}{Md~Rizwan Parvez}.} \bibinfo{year}{2024}\natexlab{}.
\newblock \showarticletitle{MapCoder: Multi-Agent Code Generation for Competitive Problem Solving}.
\newblock \bibinfo{journal}{\emph{arXiv preprint arXiv:2405.11403}} (\bibinfo{year}{2024}).
\newblock


\bibitem[Jang et~al\mbox{.}(2024)]%
        {jang2024eden}
\bibfield{author}{\bibinfo{person}{Jihoon Jang}, \bibinfo{person}{Hyokeun Lee}, {and} \bibinfo{person}{Hyun Kim}.} \bibinfo{year}{2024}\natexlab{}.
\newblock \showarticletitle{EDeN: Enabling Low-Power CNN Inference on Edge Devices Using Prefetcher-assisted NVM Systems}. In \bibinfo{booktitle}{\emph{Proceedings of the 29th ACM/IEEE International Symposium on Low Power Electronics and Design}}. \bibinfo{pages}{1--6}.
\newblock


\bibitem[Jeon et~al\mbox{.}(2025)]%
        {jeon2025tinymem}
\bibfield{author}{\bibinfo{person}{Changmin Jeon}, \bibinfo{person}{Taesik Gong}, \bibinfo{person}{Juheon Yi}, \bibinfo{person}{Fahim Kawsar}, {and} \bibinfo{person}{Chulhong Min}.} \bibinfo{year}{2025}\natexlab{}.
\newblock \showarticletitle{TinyMem: Boosting Multi-DNN Inference on Tiny AI Accelerators with Weight Memory Virtualization}. In \bibinfo{booktitle}{\emph{Proceedings of the 26th International Workshop on Mobile Computing Systems and Applications}}. \bibinfo{pages}{1--6}.
\newblock


\bibitem[Jia et~al\mbox{.}(2024)]%
        {jia2024tinytta}
\bibfield{author}{\bibinfo{person}{Hong Jia}, \bibinfo{person}{Young Kwon}, \bibinfo{person}{Alessio Orsino}, \bibinfo{person}{Ting Dang}, \bibinfo{person}{Domenico Talia}, {and} \bibinfo{person}{Cecilia Mascolo}.} \bibinfo{year}{2024}\natexlab{}.
\newblock \showarticletitle{TinyTTA: Efficient Test-time Adaptation via Early-exit Ensembles on Edge Devices}.
\newblock \bibinfo{journal}{\emph{Advances in Neural Information Processing Systems}}  \bibinfo{volume}{37} (\bibinfo{year}{2024}), \bibinfo{pages}{43274--43299}.
\newblock


\bibitem[Jin et~al\mbox{.}(2020)]%
        {jin2020compiling}
\bibfield{author}{\bibinfo{person}{Tian Jin}, \bibinfo{person}{Gheorghe-Teodor Bercea}, \bibinfo{person}{Tung~D Le}, \bibinfo{person}{Tong Chen}, \bibinfo{person}{Gong Su}, \bibinfo{person}{Haruki Imai}, \bibinfo{person}{Yasushi Negishi}, \bibinfo{person}{Anh Leu}, \bibinfo{person}{Kevin O'Brien}, \bibinfo{person}{Kiyokuni Kawachiya}, {et~al\mbox{.}}} \bibinfo{year}{2020}\natexlab{}.
\newblock \showarticletitle{Compiling onnx neural network models using mlir}.
\newblock \bibinfo{journal}{\emph{arXiv preprint arXiv:2008.08272}} (\bibinfo{year}{2020}).
\newblock


\bibitem[Kwon et~al\mbox{.}(2022)]%
        {kwon2022fast}
\bibfield{author}{\bibinfo{person}{Woosuk Kwon}, \bibinfo{person}{Sehoon Kim}, \bibinfo{person}{Michael~W Mahoney}, \bibinfo{person}{Joseph Hassoun}, \bibinfo{person}{Kurt Keutzer}, {and} \bibinfo{person}{Amir Gholami}.} \bibinfo{year}{2022}\natexlab{}.
\newblock \showarticletitle{A fast post-training pruning framework for transformers}.
\newblock \bibinfo{journal}{\emph{Advances in Neural Information Processing Systems}}  \bibinfo{volume}{35} (\bibinfo{year}{2022}), \bibinfo{pages}{24101--24116}.
\newblock


\bibitem[Lee and Yannakakis(1996)]%
        {lee1996principles}
\bibfield{author}{\bibinfo{person}{David Lee} {and} \bibinfo{person}{Mihalis Yannakakis}.} \bibinfo{year}{1996}\natexlab{}.
\newblock \showarticletitle{Principles and methods of testing finite state machines-a survey}.
\newblock \bibinfo{journal}{\emph{Proc. IEEE}} \bibinfo{volume}{84}, \bibinfo{number}{8} (\bibinfo{year}{1996}), \bibinfo{pages}{1090--1123}.
\newblock


\bibitem[Li et~al\mbox{.}(2021)]%
        {li2021hermes}
\bibfield{author}{\bibinfo{person}{Ang Li}, \bibinfo{person}{Jingwei Sun}, \bibinfo{person}{Pengcheng Li}, \bibinfo{person}{Yu Pu}, \bibinfo{person}{Hai Li}, {and} \bibinfo{person}{Yiran Chen}.} \bibinfo{year}{2021}\natexlab{}.
\newblock \showarticletitle{Hermes: an efficient federated learning framework for heterogeneous mobile clients}. In \bibinfo{booktitle}{\emph{{ACM} {MobiCom}}}. \bibinfo{pages}{420--437}.
\newblock


\bibitem[Li et~al\mbox{.}(2026)]%
        {li2026autoran}
\bibfield{author}{\bibinfo{person}{Xin Li}, \bibinfo{person}{Shiming Yu}, \bibinfo{person}{Leming Shen}, \bibinfo{person}{Jianing Zhang}, \bibinfo{person}{Yuanqing Zheng}, {and} \bibinfo{person}{Yaxiong Xie}.} \bibinfo{year}{2026}\natexlab{}.
\newblock \showarticletitle{AutORAN: LLM-driven Natural Language Programming for Agile xApp Development}.
\newblock \bibinfo{journal}{\emph{arXiv preprint arXiv:2603.18604}} (\bibinfo{year}{2026}).
\newblock


\bibitem[Liu et~al\mbox{.}(2024b)]%
        {liu2024dynaspa}
\bibfield{author}{\bibinfo{person}{Renyuan Liu}, \bibinfo{person}{Yuyang Leng}, \bibinfo{person}{Shilei Tian}, \bibinfo{person}{Shaohan Hu}, \bibinfo{person}{Chun-Fu Chen}, {and} \bibinfo{person}{Shuochao Yao}.} \bibinfo{year}{2024}\natexlab{b}.
\newblock \showarticletitle{DynaSpa: Exploiting Spatial Sparsity for Efficient Dynamic DNN Inference on Devices}. In \bibinfo{booktitle}{\emph{Proceedings of the 22nd ACM Conference on Embedded Networked Sensor Systems}}. \bibinfo{pages}{422--435}.
\newblock


\bibitem[Liu et~al\mbox{.}(2024a)]%
        {liu2024mmbench}
\bibfield{author}{\bibinfo{person}{Yuan Liu}, \bibinfo{person}{Haodong Duan}, \bibinfo{person}{Yuanhan Zhang}, \bibinfo{person}{Bo Li}, \bibinfo{person}{Songyang Zhang}, \bibinfo{person}{Wangbo Zhao}, \bibinfo{person}{Yike Yuan}, \bibinfo{person}{Jiaqi Wang}, \bibinfo{person}{Conghui He}, \bibinfo{person}{Ziwei Liu}, {et~al\mbox{.}}} \bibinfo{year}{2024}\natexlab{a}.
\newblock \showarticletitle{Mmbench: Is your multi-modal model an all-around player?}. In \bibinfo{booktitle}{\emph{European conference on computer vision}}. Springer, \bibinfo{pages}{216--233}.
\newblock


\bibitem[Liu et~al\mbox{.}(2018)]%
        {liu2018rethinking}
\bibfield{author}{\bibinfo{person}{Zhuang Liu}, \bibinfo{person}{Mingjie Sun}, \bibinfo{person}{Tinghui Zhou}, \bibinfo{person}{Gao Huang}, {and} \bibinfo{person}{Trevor Darrell}.} \bibinfo{year}{2018}\natexlab{}.
\newblock \showarticletitle{Rethinking the value of network pruning}.
\newblock \bibinfo{journal}{\emph{arXiv preprint arXiv:1810.05270}} (\bibinfo{year}{2018}).
\newblock


\bibitem[Lu et~al\mbox{.}(2024)]%
        {lu2024deepseek}
\bibfield{author}{\bibinfo{person}{Haoyu Lu}, \bibinfo{person}{Wen Liu}, \bibinfo{person}{Bo Zhang}, \bibinfo{person}{Bingxuan Wang}, \bibinfo{person}{Kai Dong}, \bibinfo{person}{Bo Liu}, \bibinfo{person}{Jingxiang Sun}, \bibinfo{person}{Tongzheng Ren}, \bibinfo{person}{Zhuoshu Li}, \bibinfo{person}{Hao Yang}, {et~al\mbox{.}}} \bibinfo{year}{2024}\natexlab{}.
\newblock \showarticletitle{Deepseek-vl: towards real-world vision-language understanding}.
\newblock \bibinfo{journal}{\emph{arXiv preprint arXiv:2403.05525}} (\bibinfo{year}{2024}).
\newblock


\bibitem[Luo et~al\mbox{.}(2024)]%
        {luo2024repoagent}
\bibfield{author}{\bibinfo{person}{Qinyu Luo}, \bibinfo{person}{Yining Ye}, \bibinfo{person}{Shihao Liang}, \bibinfo{person}{Zhong Zhang}, \bibinfo{person}{Yujia Qin}, \bibinfo{person}{Yaxi Lu}, \bibinfo{person}{Yesai Wu}, \bibinfo{person}{Xin Cong}, \bibinfo{person}{Yankai Lin}, \bibinfo{person}{Yingli Zhang}, {et~al\mbox{.}}} \bibinfo{year}{2024}\natexlab{}.
\newblock \showarticletitle{Repoagent: An llm-powered open-source framework for repository-level code documentation generation}.
\newblock \bibinfo{journal}{\emph{arXiv preprint arXiv:2402.16667}} (\bibinfo{year}{2024}).
\newblock


\bibitem[Merkel(2014)]%
        {docker}
\bibfield{author}{\bibinfo{person}{Dirk Merkel}.} \bibinfo{year}{2014}\natexlab{}.
\newblock \showarticletitle{Docker: lightweight Linux containers for consistent development and deployment}.
\newblock \bibinfo{journal}{\emph{Linux J.}} (\bibinfo{year}{2014}).
\newblock


\bibitem[Merkel et~al\mbox{.}(2014)]%
        {merkel2014docker}
\bibfield{author}{\bibinfo{person}{Dirk Merkel} {et~al\mbox{.}}} \bibinfo{year}{2014}\natexlab{}.
\newblock \showarticletitle{Docker: lightweight linux containers for consistent development and deployment}.
\newblock \bibinfo{journal}{\emph{Linux j}} \bibinfo{volume}{239}, \bibinfo{number}{2} (\bibinfo{year}{2014}), \bibinfo{pages}{2}.
\newblock


\bibitem[NVIDIA({[n.\,d.]})]%
        {tensorrt}
\bibfield{author}{\bibinfo{person}{NVIDIA}.} \bibinfo{year}{[n.\,d.]}\natexlab{}.
\newblock \bibinfo{title}{TensorRT Open Source Software}.
\newblock \bibinfo{howpublished}{\url{https://github.com/NVIDIA/TensorRT}}.
\newblock


\bibitem[Ouyang et~al\mbox{.}(2023)]%
        {ouyang2023harmony}
\bibfield{author}{\bibinfo{person}{Xiaomin Ouyang}, \bibinfo{person}{Zhiyuan Xie}, \bibinfo{person}{Heming Fu}, \bibinfo{person}{Sitong Cheng}, \bibinfo{person}{Li Pan}, \bibinfo{person}{Neiwen Ling}, \bibinfo{person}{Guoliang Xing}, \bibinfo{person}{Jiayu Zhou}, {and} \bibinfo{person}{Jianwei Huang}.} \bibinfo{year}{2023}\natexlab{}.
\newblock \showarticletitle{Harmony: Heterogeneous multi-modal federated learning through disentangled model training}. In \bibinfo{booktitle}{\emph{{ACM} {MobiSys}}}. \bibinfo{pages}{530--543}.
\newblock


\bibitem[Park and Ko(2024)]%
        {park2024fedhm}
\bibfield{author}{\bibinfo{person}{JaeYeon Park} {and} \bibinfo{person}{JeongGil Ko}.} \bibinfo{year}{2024}\natexlab{}.
\newblock \showarticletitle{FedHM: Practical federated learning for heterogeneous model deployments}.
\newblock \bibinfo{journal}{\emph{ICT Express}} \bibinfo{volume}{10}, \bibinfo{number}{2} (\bibinfo{year}{2024}), \bibinfo{pages}{387--392}.
\newblock


\bibitem[Paszke et~al\mbox{.}(2019)]%
        {paszke2019pytorch}
\bibfield{author}{\bibinfo{person}{Adam Paszke}, \bibinfo{person}{Sam Gross}, \bibinfo{person}{Francisco Massa}, \bibinfo{person}{Adam Lerer}, \bibinfo{person}{James Bradbury}, \bibinfo{person}{Gregory Chanan}, \bibinfo{person}{Trevor Killeen}, \bibinfo{person}{Zeming Lin}, \bibinfo{person}{Natalia Gimelshein}, \bibinfo{person}{Luca Antiga}, {et~al\mbox{.}}} \bibinfo{year}{2019}\natexlab{}.
\newblock \showarticletitle{Pytorch: An imperative style, high-performance deep learning library}.
\newblock \bibinfo{journal}{\emph{Advances in neural information processing systems}}  \bibinfo{volume}{32} (\bibinfo{year}{2019}).
\newblock


\bibitem[Pichora-Fuller and Dupuis(2020)]%
        {SP2/E8H2MF_2020}
\bibfield{author}{\bibinfo{person}{M.~Kathleen Pichora-Fuller} {and} \bibinfo{person}{Kate Dupuis}.} \bibinfo{year}{2020}\natexlab{}.
\newblock \bibinfo{title}{{Toronto emotional speech set (TESS)}}.
\newblock


\bibitem[Pochelu(2022)]%
        {pochelu2022deep}
\bibfield{author}{\bibinfo{person}{Pierrick Pochelu}.} \bibinfo{year}{2022}\natexlab{}.
\newblock \showarticletitle{Deep Learning Inference Frameworks Benchmark}.
\newblock \bibinfo{journal}{\emph{arXiv preprint arXiv:2210.04323}} (\bibinfo{year}{2022}).
\newblock


\bibitem[Redmon et~al\mbox{.}(2016)]%
        {redmon2016you}
\bibfield{author}{\bibinfo{person}{Joseph Redmon}, \bibinfo{person}{Santosh Divvala}, \bibinfo{person}{Ross Girshick}, {and} \bibinfo{person}{Ali Farhadi}.} \bibinfo{year}{2016}\natexlab{}.
\newblock \showarticletitle{You only look once: Unified, real-time object detection}. In \bibinfo{booktitle}{\emph{{IEEE} {CVPR}}}. \bibinfo{pages}{779--788}.
\newblock


\bibitem[Shen et~al\mbox{.}(2026a)]%
        {shen2026seeing}
\bibfield{author}{\bibinfo{person}{Leming Shen}, \bibinfo{person}{Shikai Geng}, \bibinfo{person}{Yuanqing Zheng}, {and} \bibinfo{person}{Chris~Xiaoxuan Lu}.} \bibinfo{year}{2026}\natexlab{a}.
\newblock \showarticletitle{Seeing is Not Believing: Breaking the Physical-to-Digital Trust Boundary in Robotics}.
\newblock \bibinfo{journal}{\emph{arXiv preprint arXiv:2609.08280}} (\bibinfo{year}{2026}).
\newblock


\bibitem[Shen et~al\mbox{.}(2024)]%
        {shen2024fedconv}
\bibfield{author}{\bibinfo{person}{Leming Shen}, \bibinfo{person}{Qiang Yang}, \bibinfo{person}{Kaiyan Cui}, \bibinfo{person}{Yuanqing Zheng}, \bibinfo{person}{Xiao-Yong Wei}, \bibinfo{person}{Jianwei Liu}, {and} \bibinfo{person}{Jinsong Han}.} \bibinfo{year}{2024}\natexlab{}.
\newblock \showarticletitle{FedConv: A Learning-on-Model Paradigm for Heterogeneous Federated Clients}. In \bibinfo{booktitle}{\emph{{ACM} {MobiSys}}}. \bibinfo{pages}{398--411}.
\newblock


\bibitem[Shen et~al\mbox{.}(2025a)]%
        {shen2025hierarchical}
\bibfield{author}{\bibinfo{person}{Leming Shen}, \bibinfo{person}{Qiang Yang}, \bibinfo{person}{Kaiyan Cui}, \bibinfo{person}{Yuanqing Zheng}, \bibinfo{person}{Xiao-Yong Wei}, \bibinfo{person}{Jianwei Liu}, {and} \bibinfo{person}{Jinsong Han}.} \bibinfo{year}{2025}\natexlab{a}.
\newblock \showarticletitle{Hierarchical and heterogeneous federated learning via a learning-on-model paradigm}.
\newblock \bibinfo{journal}{\emph{IEEE Transactions on Mobile Computing}} (\bibinfo{year}{2025}).
\newblock


\bibitem[Shen et~al\mbox{.}(2025b)]%
        {shen2025gpiot}
\bibfield{author}{\bibinfo{person}{Leming Shen}, \bibinfo{person}{Qiang Yang}, \bibinfo{person}{Xinyu Huang}, \bibinfo{person}{Zijing Ma}, {and} \bibinfo{person}{Yuanqing Zheng}.} \bibinfo{year}{2025}\natexlab{b}.
\newblock \showarticletitle{GPIoT: Tailoring Small Language Models for IoT Program Synthesis and Development}. In \bibinfo{booktitle}{\emph{{ACM} {SenSys}}}.
\newblock


\bibitem[Shen et~al\mbox{.}(2026b)]%
        {shen2026automating}
\bibfield{author}{\bibinfo{person}{Leming Shen}, \bibinfo{person}{Qiang Yang}, \bibinfo{person}{Xinyu Huang}, \bibinfo{person}{Zijing Ma}, \bibinfo{person}{Yuanqing Zheng}, {and} \bibinfo{person}{Chris~Xiaoxuan Lu}.} \bibinfo{year}{2026}\natexlab{b}.
\newblock \showarticletitle{Automating Model Deployment on Mobile Devices}. In \bibinfo{booktitle}{\emph{Proceedings of the 24th Annual International Conference on Mobile Systems, Applications and Services Workshops}}. \bibinfo{pages}{188--194}.
\newblock


\bibitem[Shen et~al\mbox{.}(2026c)]%
        {shen2026one}
\bibfield{author}{\bibinfo{person}{Leming Shen}, \bibinfo{person}{Qiang Yang}, \bibinfo{person}{Xinyu Huang}, \bibinfo{person}{Zijing Ma}, \bibinfo{person}{Yuanqing Zheng}, {and} \bibinfo{person}{Chris Xiaoxuan~Lu}.} \bibinfo{year}{2026}\natexlab{c}.
\newblock \showarticletitle{One Workflow Doesn't Fit All: Adaptive Workflows for Edge AI Development}.
\newblock \bibinfo{journal}{\emph{GetMobile: Mobile Computing and Communications}} \bibinfo{volume}{30}, \bibinfo{number}{1} (\bibinfo{year}{2026}), \bibinfo{pages}{11--15}.
\newblock


\bibitem[Shen et~al\mbox{.}(2025c)]%
        {shen2025autoiot}
\bibfield{author}{\bibinfo{person}{Leming Shen}, \bibinfo{person}{Qiang Yang}, \bibinfo{person}{Yuanqing Zheng}, {and} \bibinfo{person}{Mo Li}.} \bibinfo{year}{2025}\natexlab{c}.
\newblock \showarticletitle{AutoIOT: LLM-Driven Automated Natural Language Programming for AIoT Applications}. In \bibinfo{booktitle}{\emph{{ACM} {MobiCom}}}.
\newblock


\bibitem[Shen and Zheng(2024)]%
        {shen2024iotcoder}
\bibfield{author}{\bibinfo{person}{Leming Shen} {and} \bibinfo{person}{Yuanqing Zheng}.} \bibinfo{year}{2024}\natexlab{}.
\newblock \showarticletitle{Iotcoder: A copilot for iot application development}. In \bibinfo{booktitle}{\emph{Proceedings of the 30th Annual International Conference on Mobile Computing and Networking}}. \bibinfo{pages}{1647--1649}.
\newblock


\bibitem[Shinn et~al\mbox{.}(2023)]%
        {shinn2023reflexion}
\bibfield{author}{\bibinfo{person}{Noah Shinn}, \bibinfo{person}{Federico Cassano}, \bibinfo{person}{Ashwin Gopinath}, \bibinfo{person}{Karthik Narasimhan}, {and} \bibinfo{person}{Shunyu Yao}.} \bibinfo{year}{2023}\natexlab{}.
\newblock \showarticletitle{Reflexion: Language agents with verbal reinforcement learning}.
\newblock \bibinfo{journal}{\emph{Advances in Neural Information Processing Systems}}  \bibinfo{volume}{36} (\bibinfo{year}{2023}), \bibinfo{pages}{8634--8652}.
\newblock


\bibitem[Wei et~al\mbox{.}(2022)]%
        {wei2022chain}
\bibfield{author}{\bibinfo{person}{Jason Wei}, \bibinfo{person}{Xuezhi Wang}, \bibinfo{person}{Dale Schuurmans}, \bibinfo{person}{Maarten Bosma}, \bibinfo{person}{Fei Xia}, \bibinfo{person}{Ed Chi}, \bibinfo{person}{Quoc~V Le}, \bibinfo{person}{Denny Zhou}, {et~al\mbox{.}}} \bibinfo{year}{2022}\natexlab{}.
\newblock \showarticletitle{Chain-of-thought prompting elicits reasoning in large language models}.
\newblock \bibinfo{journal}{\emph{Advances in neural information processing systems}}  \bibinfo{volume}{35} (\bibinfo{year}{2022}), \bibinfo{pages}{24824--24837}.
\newblock


\bibitem[Weng(2021)]%
        {weng2021neural}
\bibfield{author}{\bibinfo{person}{Olivia Weng}.} \bibinfo{year}{2021}\natexlab{}.
\newblock \showarticletitle{Neural network quantization for efficient inference: A survey}.
\newblock \bibinfo{journal}{\emph{arXiv preprint arXiv:2112.06126}} (\bibinfo{year}{2021}).
\newblock


\bibitem[Xu et~al\mbox{.}(2024a)]%
        {xu2024penetrative}
\bibfield{author}{\bibinfo{person}{Huatao Xu}, \bibinfo{person}{Liying Han}, \bibinfo{person}{Qirui Yang}, \bibinfo{person}{Mo Li}, {and} \bibinfo{person}{Mani Srivastava}.} \bibinfo{year}{2024}\natexlab{a}.
\newblock \showarticletitle{Penetrative ai: Making llms comprehend the physical world}. In \bibinfo{booktitle}{\emph{{ACM} {HOTMOBILE}}}. \bibinfo{pages}{1--7}.
\newblock


\bibitem[Xu et~al\mbox{.}(2024b)]%
        {xu2024can}
\bibfield{author}{\bibinfo{person}{Zhenyu Xu}, \bibinfo{person}{Hailin Xu}, \bibinfo{person}{Zhouyang Lu}, \bibinfo{person}{Yingying Zhao}, \bibinfo{person}{Rui Zhu}, \bibinfo{person}{Yujiang Wang}, \bibinfo{person}{Mingzhi Dong}, \bibinfo{person}{Yuhu Chang}, \bibinfo{person}{Qin Lv}, \bibinfo{person}{Robert~P Dick}, {et~al\mbox{.}}} \bibinfo{year}{2024}\natexlab{b}.
\newblock \showarticletitle{Can large language models be good companions? An LLM-based eyewear system with conversational common ground}.
\newblock \bibinfo{journal}{\emph{{ACM} {IMWUT}}} \bibinfo{volume}{8}, \bibinfo{number}{2} (\bibinfo{year}{2024}), \bibinfo{pages}{1--41}.
\newblock


\bibitem[Yang et~al\mbox{.}(2023)]%
        {yang2023edgefm}
\bibfield{author}{\bibinfo{person}{Bufang Yang}, \bibinfo{person}{Lixing He}, \bibinfo{person}{Neiwen Ling}, \bibinfo{person}{Zhenyu Yan}, \bibinfo{person}{Guoliang Xing}, \bibinfo{person}{Xian Shuai}, \bibinfo{person}{Xiaozhe Ren}, {and} \bibinfo{person}{Xin Jiang}.} \bibinfo{year}{2023}\natexlab{}.
\newblock \showarticletitle{Edgefm: Leveraging foundation model for open-set learning on the edge}. In \bibinfo{booktitle}{\emph{{ACM} {SenSys}}}.
\newblock


\bibitem[Yang and Shami(2020)]%
        {yang2020hyperparameter}
\bibfield{author}{\bibinfo{person}{Li Yang} {and} \bibinfo{person}{Abdallah Shami}.} \bibinfo{year}{2020}\natexlab{}.
\newblock \showarticletitle{On hyperparameter optimization of machine learning algorithms: Theory and practice}.
\newblock \bibinfo{journal}{\emph{Neurocomputing}}  \bibinfo{volume}{415} (\bibinfo{year}{2020}), \bibinfo{pages}{295--316}.
\newblock


\bibitem[Yao et~al\mbox{.}(2023a)]%
        {yao2023tree}
\bibfield{author}{\bibinfo{person}{Shunyu Yao}, \bibinfo{person}{Dian Yu}, \bibinfo{person}{Jeffrey Zhao}, \bibinfo{person}{Izhak Shafran}, \bibinfo{person}{Tom Griffiths}, \bibinfo{person}{Yuan Cao}, {and} \bibinfo{person}{Karthik Narasimhan}.} \bibinfo{year}{2023}\natexlab{a}.
\newblock \showarticletitle{Tree of thoughts: Deliberate problem solving with large language models}.
\newblock \bibinfo{journal}{\emph{Advances in neural information processing systems}}  \bibinfo{volume}{36} (\bibinfo{year}{2023}), \bibinfo{pages}{11809--11822}.
\newblock


\bibitem[Yao et~al\mbox{.}(2023b)]%
        {yao2023react}
\bibfield{author}{\bibinfo{person}{Shunyu Yao}, \bibinfo{person}{Jeffrey Zhao}, \bibinfo{person}{Dian Yu}, \bibinfo{person}{Nan Du}, \bibinfo{person}{Izhak Shafran}, \bibinfo{person}{Karthik Narasimhan}, {and} \bibinfo{person}{Yuan Cao}.} \bibinfo{year}{2023}\natexlab{b}.
\newblock \showarticletitle{React: Synergizing reasoning and acting in language models}. In \bibinfo{booktitle}{\emph{International Conference on Learning Representations (ICLR)}}.
\newblock


\bibitem[Yue et~al\mbox{.}(2024)]%
        {yue2024mmmu}
\bibfield{author}{\bibinfo{person}{Xiang Yue}, \bibinfo{person}{Yuansheng Ni}, \bibinfo{person}{Kai Zhang}, \bibinfo{person}{Tianyu Zheng}, \bibinfo{person}{Ruoqi Liu}, \bibinfo{person}{Ge Zhang}, \bibinfo{person}{Samuel Stevens}, \bibinfo{person}{Dongfu Jiang}, \bibinfo{person}{Weiming Ren}, \bibinfo{person}{Yuxuan Sun}, {et~al\mbox{.}}} \bibinfo{year}{2024}\natexlab{}.
\newblock \showarticletitle{Mmmu: A massive multi-discipline multimodal understanding and reasoning benchmark for expert agi}. In \bibinfo{booktitle}{\emph{Proceedings of the IEEE/CVF Conference on Computer Vision and Pattern Recognition}}. \bibinfo{pages}{9556--9567}.
\newblock


\bibitem[Zhang et~al\mbox{.}(2021)]%
        {zhang2021nn}
\bibfield{author}{\bibinfo{person}{Li~Lyna Zhang}, \bibinfo{person}{Shihao Han}, \bibinfo{person}{Jianyu Wei}, \bibinfo{person}{Ningxin Zheng}, \bibinfo{person}{Ting Cao}, \bibinfo{person}{Yuqing Yang}, {and} \bibinfo{person}{Yunxin Liu}.} \bibinfo{year}{2021}\natexlab{}.
\newblock \showarticletitle{Nn-meter: Towards accurate latency prediction of deep-learning model inference on diverse edge devices}. In \bibinfo{booktitle}{\emph{{ACM} {MobiSys}}}. \bibinfo{pages}{81--93}.
\newblock


\bibitem[Zhang et~al\mbox{.}(2025)]%
        {zhang2025vlm}
\bibfield{author}{\bibinfo{person}{Zezhong Zhang}, \bibinfo{person}{Chenyu Hu}, \bibinfo{person}{Sunwoh Lye}, {and} \bibinfo{person}{Chen Lv}.} \bibinfo{year}{2025}\natexlab{}.
\newblock \showarticletitle{A VLM-Drone System for Indoor Navigation Assistance with Semantic Reasoning for the Visually Impaired}. In \bibinfo{booktitle}{\emph{2025 IEEE/SICE International Symposium on System Integration (SII)}}. IEEE, \bibinfo{pages}{1260--1265}.
\newblock


\end{thebibliography}

\end{sloppypar}
\end{document}